\documentclass{article}
\usepackage{crosstimeedit_preprint,times}
\usepackage[T1]{fontenc}
\usepackage[utf8]{inputenc}
\usepackage{amsmath,amssymb,graphicx,booktabs,array,tabularx,makecell,multirow}
\usepackage[table]{xcolor}
\usepackage{pifont,caption,placeins,needspace,float,wrapfig}
\usepackage{framed}
\usepackage{fvextra}
\newenvironment{promptbox}{\begin{framed}}{\end{framed}}
\RecustomVerbatimEnvironment{verbatim}{Verbatim}{breaklines=true,breakanywhere=true,breaksymbolleft={},breaksymbolright={}}
\usepackage{algorithm,algorithmic}

\newcolumntype{Y}{>{\centering\arraybackslash}X}
\DeclareCaptionFont{eightpt}{\fontsize{8}{9}\selectfont}

\usepackage{hyperref,url}
\hypersetup{colorlinks=true,linkcolor=blue,citecolor=blue,urlcolor=blue,pdfauthor={Hanwen Lu, Jun He, Mingjia Yang, Hao Wei, Jinhao Huang, Yi Lin, Xiang Zhang},pdftitle={CrossTimeEdit: A Decade-Spanning Cross-View Dataset and Reward-Guided Editing for Historical Street-View Generation},pdfsubject={Preprint}}
\newcommand{\yes}{\ding{51}}
\newcommand{\no}{\ding{55}}

\definecolor{oursbg}{RGB}{237,235,255}
\title{\raggedright CrossTimeEdit: A Decade-Spanning Cross-View Dataset and Reward-Guided Editing for Historical Street-View Generation}
\author{
  Hanwen Lu\textsuperscript{1}\thanks{
    Equal contribution.\quad
    $\dagger$ Corresponding author. E-mail:
    \texttt{zhangx795@mail.sysu.edu.cn}.
  },
  Jun He\textsuperscript{1}\footnotemark[\value{footnote}],
  Mingjia Yang\textsuperscript{1},
  Hao Wei\textsuperscript{1},
  Jinhao Huang\textsuperscript{1},
  Yi Lin\textsuperscript{1}, \\
  \textbf{
  Xiang Zhang\textsuperscript{1}$^{\dagger}$
  } \\
  \textsuperscript{1}Sun Yat-sen University
}
\begin{document}
\raggedbottom 
\maketitle
\begin{center}
\includegraphics[width=\linewidth]{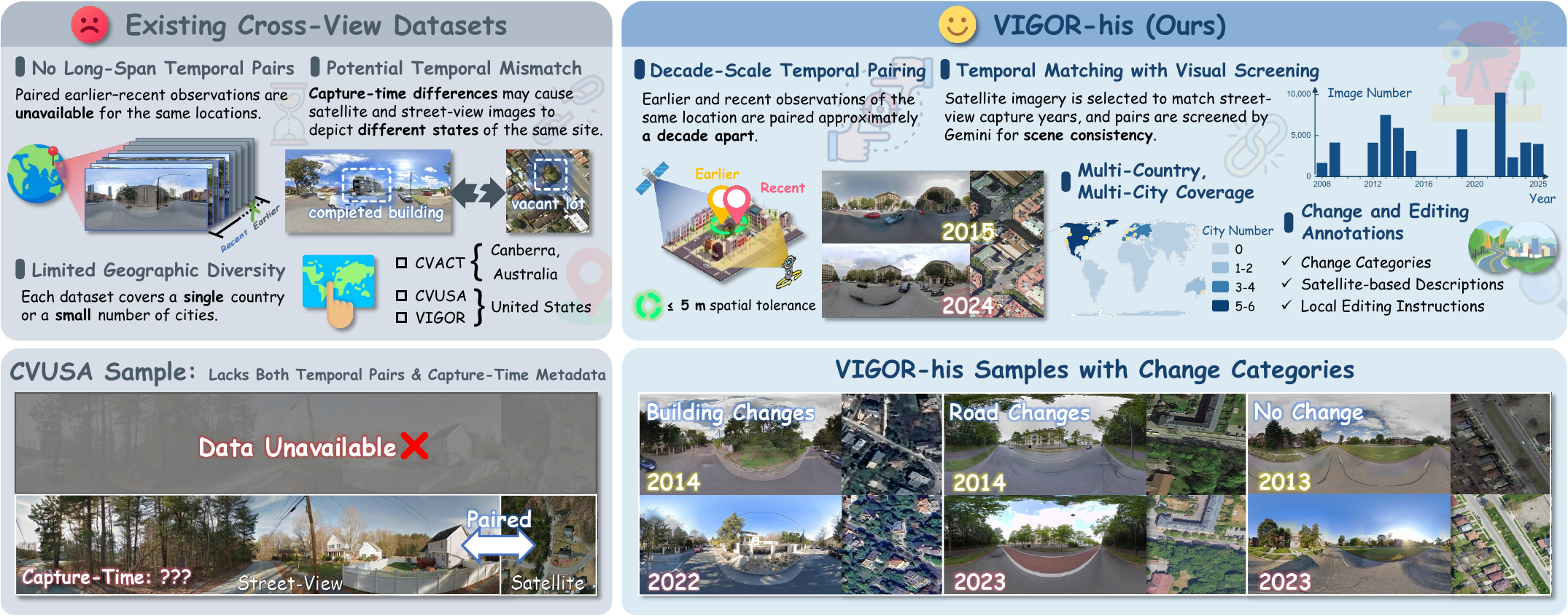}
\captionof{figure}{Comparison with existing cross-view datasets. VIGOR-his dataset provides four features to our CrossTimeEdit: (1) paired earlier and recent street and satellite views spanning approximately a decade; (2) acquisition-period matching and Gemini-based screening for cross-view scene consistency; (3) geographic coverage across 11 cities on three continents; and (4) change categories, satellite-based change descriptions, and local editing instructions, supporting historical street-view generation from cross-view change evidence.}
\label{fig:teaser}
\end{center}
\begin{abstract}
Historical street-view imagery records urban evolution, but uneven coverage leaves substantial gaps in historical records. Generating plausible past appearances requires restoring changed structures while preserving persistent scene content. We construct VIGOR-his, a decade-spanning cross-view dataset containing 43,653 location-level quadruplets across 11 cities on three continents. Its automated pipeline performs spatial pairing, consistency screening, change classification, and the generation and validation of satellite-based change descriptions and local editing instructions. Based on VIGOR-his, we propose CrossTimeEdit, a model that reformulates historical street-view generation as editing, using recent street views to constrain viewpoint and unchanged appearance and temporal satellite differences as change evidence. Starting from FLUX.2 [Klein] 4B, we train CrossTimeEdit through supervised fine-tuning (SFT) followed by online reinforcement learning (RL). We design three street-view editing criteria, namely Instruction Alignment (IA), Background Preservation (BP), and Quality and Physical Plausibility (QP), as both RL reward dimensions and evaluation metrics. We optimize this multi-reward objective using Within Group Relative Policy Optimization for flow-matching models (Flow-GRPO) with Group reward-Decoupled Normalization Policy Optimization (GDPO), which normalizes each reward dimension before aggregation. CrossTimeEdit improves overall performance across the three editing criteria by 17.12\% over the pretrained baseline and outperforms cross-view generation models in scene consistency, visual realism, and perceptual quality. The implementation code, dataset, and model weights are available at \url{https://luhanwen67.github.io/CrossTimeEdit-release/}.
\end{abstract}

\section{Introduction}
Historical street-view imagery records urban evolution from a human perspective, revealing building facades, road configurations, and streetscape details that overhead imagery cannot fully capture \citep{biljecki2021svi}. Repeated observations support fine-grained analysis of redevelopment and changes in everyday urban environments \citep{r5}, complementing satellite-based assessments with street-level evidence. However, uneven spatial coverage and irregular acquisition limit comparisons across locations and periods \citep{li2022svi}. Recent images cannot show buildings that have disappeared or roads before reconfiguration. Historical street-view generation seeks to recover plausible past appearances where direct observations are unavailable.

Satellite-to-street generation offers a possible path for recovering missing historical street views from satellite-derived conditions. Cross-view generation models, including Sat2Density~\citep{r7}, Sat2Scene~\citep{r8}, ControlS2S~\citep{r9}, CrossViewDiff~\citep{r23}, and SatDreamer360~\citep{r6}, improve cross-view geometry or diffusion control, but still generate complete street views from random noise. Because satellite imagery sparsely constrains facades, textures, and occluded structures, such generation may alter buildings and roads that persisted across time, resulting in lower-quality historical street-view outputs.

These limitations motivate local editing. Earlier and recent street views at the same location share viewpoint and persistent content, whereas noise-based generation may discard texture and semantic information and fail to preserve unchanged regions. Using the recent street view as a visual reference anchors viewpoint (location and height) and unchanged appearance, while temporal satellite differences specify the earlier states of changed regions. General image editing models such as InstructPix2Pix~\citep{r12}, MagicBrush~\citep{magicbrush}, and OmniGen2~\citep{r19} may still be unsuitable for location-specific recovery in distorted panoramic street-view images, motivating a street-view-specific editing model.

This formulation requires temporal street--satellite pairs and instructions linking satellite-level changes to ground-level editing operations. Existing datasets provide only parts of this requirement: VIGOR~\citep{r3} lacks paired earlier and recent observations, CityPulse~\citep{r5} lacks corresponding satellite pairs, and CVUSA~\citep{workman2015localize} lacks temporal observations. Thus, temporal cross-view evidence and editing supervision are not jointly available for historical street-view recovery.

To address these gaps, we construct VIGOR-his with an automated pipeline for spatial pairing, consistency screening, change classification, satellite-based change description, local instruction generation, and instruction validation. The dataset provides decade-spanning cross-view quadruplets, satellite-based change descriptions, and satellite-grounded local editing instructions. We then train CrossTimeEdit, a street-view editing model, using VIGOR-his, starting from FLUX.2 [Klein] 4B and applying supervised fine-tuning (SFT) followed by online reinforcement learning (RL). We design three criteria tailored to street-view editing: Instruction Alignment (IA), Background Preservation (BP), and Quality and Physical Plausibility (QP). These criteria jointly provide the three reward dimensions for RL and the evaluation protocol for street-view editing. To optimize this multi-reward objective, we adopt Within Group Relative Policy Optimization for flow-matching models (Flow-GRPO), an online RL framework for flow-matching generation, and incorporate Group reward-Decoupled Normalization Policy Optimization (GDPO) to normalize each reward dimension independently before aggregation, accounting for their unequal variability.

\begin{itemize}
\item We introduce \textbf{VIGOR-his}: 43,653 decade-spanning cross-view quadruplets across 11 cities on three continents, with change categories, satellite-based change descriptions, and satellite-grounded local editing instructions.
\item We design a \textbf{three-criterion VLM reward and evaluation protocol} based on Instruction Alignment (IA), Background Preservation (BP), and Quality and Physical Plausibility (QP), combining earlier-target edit verification, recent-input preservation assessment, and panorama-aware quality checks for street-view editing.
\item We develop \textbf{CrossTimeEdit} through SFT and GDPO-normalized Flow-GRPO, improving overall performance by 17.12\% over the pretrained backbone and leading the evaluated open-source models in both editing and cross-view perceptual comparisons.
\end{itemize}

\section{Related Work}
\subsection{Cross-view and temporal street-view datasets}
CVUSA~\citep{workman2015localize}, CVACT~\citep{r2}, and VIGOR~\citep{r3} provide ground--overhead correspondences; VIGOR accommodates non-bijective matching. Retrieval methods address viewpoint and orientation gaps~\citep{cvmnet,safa,dsm}, while CVGlobal~\citep{r4} examines cross-period retrieval. Mapillary SLS~\citep{msls} and GSV-Cities~\citep{gsvcities} capture diverse street-level conditions over time, VIGOR++~\citep{r6} extends cross-view data to continuous panoramic sequences, and CV-Cities~\citep{cvcities} broadens geographic coverage. CityPulse~\citep{r5} provides street-view time series and change labels without corresponding satellite pairs. However, these resources primarily support localization, retrieval, or temporal street-view analysis and do not jointly provide location-matched earlier and recent street--satellite views with local editing instructions.

\subsection{Image editing and generative post-training}
Image translation and diffusion backbones underpin image editing~\citep{pix2pix,cyclegan,ldm,dit}. SDEdit~\citep{r11}, Prompt-to-Prompt~\citep{prompttoprompt}, InstructPix2Pix~\citep{r12}, and MagicBrush~\citep{magicbrush} develop input-conditioned or instruction-based editing; ControlNet~\citep{r13} adds spatial conditions. Diffusion-DPO~\citep{diffusiondpo}, ImageReward~\citep{imagereward}, and PickScore~\citep{pickscore} study preference optimization, while TIFA~\citep{tifa} and GenEval~\citep{geneval} examine image--text alignment. Flow-GRPO~\citep{r14,grpo} extends group-relative optimization to flow models, while GDPO~\citep{r15} separates reward dimensions before aggregation. However, these methods are generally developed for generic editing or preference optimization and may not be well suited to street-view images with substantial distortion and panoramic geometry.

\subsection{Satellite-to-street generation}
Early cross-view synthesis transferred scene layout or used conditional GANs~\citep{zhai2017,regmi2018,regmi2019}. Geometry-aware projection~\citep{lu2020,shi2022} and joint synthesis--retrieval~\citep{toker2021} strengthened overhead-to-ground alignment. Sat2Density~\citep{r7} models scene density; Sat2Scene~\citep{r8}, CrossViewDiff~\citep{r23}, ControlS2S~\citep{r9}, and SatDreamer360~\citep{r6} explore 3D or diffusion-based generation; GeoIdentity-Sat2Street~\citep{geoidentity} combines view transformation and refinement. These methods infer street appearance predominantly from satellite imagery, leaving persistent street content and temporal change localization underconstrained, which may lead to lower-quality street-view outputs.

\section{VIGOR-his}
We introduce VIGOR-his: 43,653 location-level cross-view quadruplets across 11 cities on three continents, pairing street and satellite views approximately a decade apart. Local editing instructions for changed samples support location-level urban spatio-temporal analysis and historical street-view generation. Table~\ref{tab:datasets} compares the temporal observations, text modality, and geographic coverage of existing datasets; Fig.~\ref{fig:pipeline} presents the collection, pre-processing, screening, and annotation pipeline. Examples of VIGOR-his samples are provided in Appendix~\ref{app:dataset_structure}.

\begingroup
\setlength{\intextsep}{6pt}
\begin{figure}[t]
\captionsetup{skip=4pt}
\centering\includegraphics[width=\linewidth]{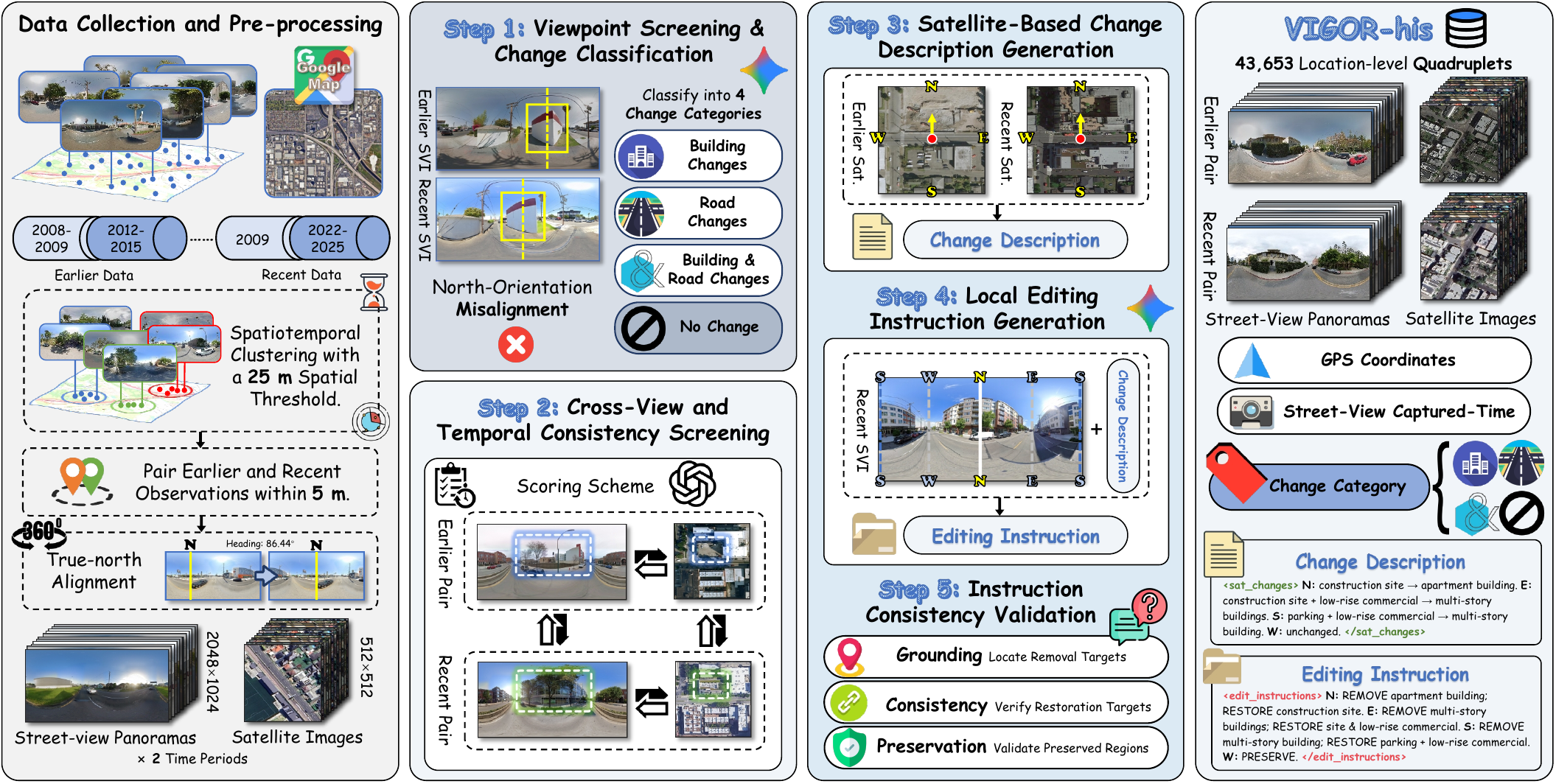}
\caption{VIGOR-his dataset construction pipeline, including data preprocessing, viewpoint screening and change classification, cross-view and temporal consistency screening, satellite-based change description generation, local editing instruction generation, and instruction consistency validation. Temporal satellite differences guide instructions describing the earlier states of changed regions, while recent street views anchor viewpoint and unchanged appearance.}
\label{fig:pipeline}
\end{figure}
\endgroup
\subsection{Data Collection and Pre-processing}
VIGOR~\citep{r3} is a cross-view geo-localization dataset with non-bijective street--satellite correspondences but no paired earlier observations. Using its panorama identifiers to locate recent observations, we manually downloaded street-view panoramas from approximately a decade earlier and corresponding satellite imagery from Google Maps\footnote{\url{https://www.google.com/maps}}. To broaden geographic coverage and data diversity, we additionally collected earlier and recent street-view and satellite images approximately a decade apart in Barcelona, Budapest, Copenhagen, Detroit, Los Angeles, London, and Sydney.

Because street-view metadata are spatially unordered, we organize observations by panorama identifier, capture time, and GPS coordinates through spatio-temporal clustering and location-level pairing. We match earlier and recent panoramas within a small spatial tolerance and center each period's satellite patch on its corresponding street-view capture location, establishing location-level temporal cross-view correspondence.

Street-view panoramas are collected at $2048\times1024$ pixels (width $\times$ height), with corresponding $512\times512$ satellite patches. Heading-based true-north alignment places north at each panorama's center, establishing a common directional reference across periods and views for comparison and local editing. Satellite patches are north-up. The complete collection and pre-processing procedure is provided in Appendix~\ref{app:pairing}.
\begin{table}[t!]
\captionsetup{font=small}
\caption{Comparison with related cross-view and temporal datasets. \yes: provided; \no: not provided in the cited release.}
\label{tab:datasets}
\centering\fontsize{8}{8}\selectfont\setlength{\tabcolsep}{0pt}\renewcommand{\arraystretch}{1.2}
\begin{tabularx}{\linewidth}{@{}l|
  >{\hsize=.80\hsize\centering\arraybackslash}X
  >{\hsize=.90\hsize\centering\arraybackslash}X
  >{\hsize=1.30\hsize\centering\arraybackslash}X
  >{\hsize=1.20\hsize\centering\arraybackslash}X
  >{\hsize=.75\hsize\centering\arraybackslash}X
  >{\hsize=1.05\hsize\centering\arraybackslash}X
  l@{}}
\toprule
\multirow{2}{*}{\textbf{Dataset}} & Cross-view & Temporal & Temporal & Cross-view & Text & \multirow{2}{*}{Continents} & \multirow{2}{*}{Size}\\
 & pairing & street views & satellite views & time alignment & modality & & \\
\midrule
CVUSA & \yes & \no & \no & \no & \no & 1 & 44,416 pairs\\
CVACT & \yes & \no & \no & \no & \no & 1 & 128,334 pairs\\
VIGOR & \yes & \no & \no & \no & \no & 1 & 105,214 panoramas\\
CVGlobal & \yes & \yes & \no & \no & \no & 6 & 134,233 pairs\\
CV-Cities & \yes & \no & \no & \no & \no & 6 & 223,736 pairs\\
VIGOR++ & \yes & \no & \no & \no & \no & 1 & 91,498 sequence pairs\\
CityPulse & \no & \yes & \no & \no & \no & 1 & 757 time series\\
\rowcolor{oursbg}\textbf{VIGOR-his} & \yes & \yes & \yes & \yes & \yes & \textbf{3} & \textbf{43,653 quadruplets}\\
\bottomrule
\end{tabularx}
\par\smallskip\raggedright\fontsize{8}{9}\selectfont
\vspace{-1\baselineskip}
\end{table}

\subsection{Change classification and consistency screening}
\textbf{Viewpoint Screening and Change Classification.} Gemini-3-Flash-Preview screens earlier--recent street-view pairs for substantial orientation mismatches using persistent building outlines and road vanishing points, while tolerating modest capture displacement. Pairs with severe heading misalignment are excluded to prevent viewpoint differences from being mistaken for temporal changes. The model then classifies each sample as building change, road change, building-and-road change, or no change. It focuses on major structural changes, such as construction, demolition, and road-network modifications, while disregarding transient objects and minor appearance variations. These labels distinguish cases suitable for local editing in historical street-view generation from no-change cases.

\textbf{Cross-View and Temporal Consistency Screening.} We use GPT-5.6-Luna as a vision-language evaluator to score cross-view consistency at both acquisition periods, comparing each street view with its corresponding satellite patch. The assessment focuses on building presence and form, road topology, and development status, while discounting shadows, transient objects, and roof details not observable from street level. For no-change samples, we add two temporal consistency checks: one compares the earlier and recent street views, and the other compares the earlier and recent satellite patches. These checks assess whether buildings, roads, and major markings remain consistent while discounting acquisition-style differences and transient objects. We retain only quadruplets for which every applicable score is at least 7 on a 0--10 scale.

\subsection{Local editing instructions and validation}
\textbf{Satellite-Grounded Change Descriptions.} Because earlier street views are unavailable at generation time, we derive change evidence from the paired satellite observations and reserve earlier street views for subsequent validation and supervision. Gemini-3.1-Pro-Preview compares the earlier and recent satellite patches, describing their states and structural changes along the four cardinal directions (S, W, N, E) relative to the capture location. The descriptions prioritize changes visible from street level and exclude roof-only modifications, shadows, and transient objects, providing object-level temporal evidence for editing.

\textbf{Local Editing Instruction Generation.} Gemini-3.1-Pro-Preview combines the satellite-based descriptions with the recent panorama, which anchors the target viewpoint, location, camera height, and unchanged appearance, to generate local editing instructions for each cardinal direction. `MODIFY' specifies a recent entity to remove and its earlier state to generate; `PRESERVE' identifies content to retain. The instructions account for occlusion, perspective, and the continuity of south-facing objects across panorama boundaries. Directions without sufficiently supported, street-visible changes are assigned `PRESERVE'; edits are localized through entities and relative positions rather than pixel masks.

\textbf{Instruction Consistency Validation.} Gemini-3.1-Flash-Lite evaluates Grounding, Consistency, and Preservation using both street views and the instructions. These checks assess whether the removal target is grounded in the images, whether the specified earlier state agrees with the earlier street view, and whether persistent changes have been omitted from `PRESERVE' regions, respectively. We retain samples scoring at least 7 on all three criteria and discard changed samples whose instructions are entirely `PRESERVE'.
\FloatBarrier

\section{CrossTimeEdit}
\subsection{Task formulation}
We consider an earlier street view $I_h$ and a recent street view $I_c$ captured at the same location, together with a local editing instruction $c$ derived from temporal satellite changes. We formulate historical street-view generation as generating an earlier street view $\widehat I_h$ from $I_c$ and $c$, where $I_h$ serves as the target reference for supervision and evaluation but is not provided as a generation input. The goal is to restore content that changed between the two periods while preserving scene content that remained unchanged.

To this end, we propose \textbf{CrossTimeEdit}. It uses recent street views as visual references and temporal satellite differences as change evidence, and realizes the task through local editing instructions derived from satellite changes: $I_c$ constrains the viewpoint and persistent appearance, while $c$ specifies the historical states of changed regions, producing $\widehat I_h$ through local image editing.

\subsection{Supervised initialization}
We train attention-projection LoRA adapters on FLUX.2 [Klein] 4B while freezing backbone weights, MLPs, and the encoder/decoder. We use $I_h$ as the target for changed examples; for no-change examples, we assign `PRESERVE' instruction to all four directions and use the input $I_c$ as the target. Following the identity-supervision principle in image editing \citep{r18}, these examples jointly teach conditional editing and content preservation through a shared flow-matching velocity-prediction objective. Subsequent RL uses changed conditions to focus on executing edits, while the preservation reward continues to constrain unrequested regions in every candidate.

\subsection{three-criterion VLM reward and evaluation protocol}
Three requirements are necessary for historical street-view editing: (1) executing requested changes, (2) retaining surrounding structures, and (3) maintaining image quality. We design a three-criterion VLM reward and evaluation protocol with Instruction Alignment (IA), Background Preservation (BP), and Quality and Physical Plausibility (QP), each scored from 0 to 10 by Gemini-3.1-Flash-Lite. The same Gemini-3.1-Flash-Lite evaluator is used for the reward and editing-evaluation protocol. The shared protocol provides online training rewards and offline editing evaluation under the same panoramic direction convention.

\textbf{IA: target-referenced edit correctness.} Given $I_c,\widehat I_h,I_h,c$, IA assesses if the edit is complete, accurate, and in the correct direction. The earlier target provides location-specific evidence beyond what the instruction can describe. Evaluation is restricted to requested operations, with global photographic style and unrelated target differences excluded. This distinguishes a plausible edited object from one consistent with the earlier state of the location.

\textbf{BP: input-referenced structural preservation.} Given $I_c,\widehat I_h,c$, BP checks semantic structures outside the requested edits. Unauthorized additions or removals of buildings, roads, walls, and fences are penalized more heavily than changes to vegetation. Weather, people, vehicles, and permitted global color/lighting shifts are not counted as structural damage. This criterion uses the recent input as the preservation reference and does not require the entire output to resemble the earlier target.

\textbf{QP: visual quality and physical plausibility.} Given $I_c,\widehat I_h,c$, QP examines if tearing, repetition, abnormal blur, implausible scale or distortion, and discontinuities at boundaries are introduced. Input defects provide the quality baseline; correctness of semantic edit is handled by IA. Together, the three criteria distinguish copying the input, excessive scene rewriting, and geometrically flawed edits, making the reward signal more informative than a single holistic judgment.

\subsection{Multi-reward Flow-GRPO}
\textbf{Online Policy Optimization.} Flow-GRPO~\citep{r14} samples candidate trajectories under the same condition using Flow-SDE, scores their terminal images, and updates the policy with a PPO-style clipped objective~\citep{ppo}. We initialize the policy from SFT and update only its LoRA parameters. The VLM evaluator provides scalar rewards for policy optimization, with no gradient propagation through the evaluator. Policy updates use transition probability ratios, while a velocity-prediction MSE penalty regularizes the policy against a frozen reference. 

\textbf{Dimension-Wise Reward Normalization.} Standard GRPO~\citep{grpo} normalizes a scalar group reward after reward aggregation. In our Flow-GRPO setting, directly aggregating the IA, BP, and QP rewards made the resulting advantage sensitive to their dimension-specific variability, allowing a high-variance dimension to dominate the update. GDPO~\citep{r15} addresses this issue by normalizing each reward dimension before aggregation. We therefore incorporate this reward-decoupled procedure into Flow-GRPO and refer to it as dimension-wise group normalization (DGN).

Let $r_{m,i}$ be candidate $i$'s score for dimension $m$, with group $g(i)$. The within-group mean and population standard deviation are denoted by $\mu_{m,g}$ and $\sigma_{m,g}$, and a positive numerical floor by $\varepsilon$. First, we normalize each reward separately:
\begin{equation}
z_{m,i}=\frac{r_{m,i}-\mu_{m,g(i)}}{\max(\sigma_{m,g(i)},\varepsilon)},\qquad m\in\{\mathrm{IA,BP,QP}\}.
\label{eq:dimnorm}
\end{equation}
Next, we aggregate the normalized rewards using empirically chosen weights fixed throughout the experiment:
\begin{equation}
a_i=\sum_m w_m z_{m,i}=0.45z_{\mathrm{IA},i}+0.30z_{\mathrm{BP},i}+0.25z_{\mathrm{QP},i}.
\label{eq:weight}
\end{equation}
Finally, we normalize the combined values across the current multi-device candidate batch $\mathcal B$, and clip the resulting advantages:
\begin{equation}
A_i=\frac{a_i-\mu_{\mathcal B}}{\max(\sigma_{\mathcal B},\varepsilon)},\qquad
\widetilde A_i=\operatorname{clip}(A_i,-A_{\max},A_{\max}).
\label{eq:batchnorm}
\end{equation}
Here, $\mu_{\mathcal B}$ and $\sigma_{\mathcal B}$ are the batch statistics of $a_i$. Dimension-wise normalization adjusts the relative scale of reward variation, while batch normalization controls the overall advantage scale. A constant reward dimension contributes zero after centering; clipping limits the influence of extreme advantages. The weights express fixed optimization preferences rather than learned coefficients.

\section{Experiments}
\subsection{Experimental setup}
\textbf{Implementation Details.} The training set contains 21,993 samples: 8,797 changed and 13,196 no-change quadruplets, approximately a 4:6 ratio. Validation and test sets contain 1,095 and 1,094 changed samples, respectively. No-change examples provide identity supervision during SFT; RL uses 2,000 randomly selected changed training conditions. Validation supports model selection and configuration choice, while the test set measures final performance.
Both stages use AdamW to train attention-projection LoRA adapters with rank 32 and $\alpha=32$. SFT comprises two five-epoch phases: the first uses a learning rate of $3\times10^{-5}$ with 412-step warmup and a constant schedule, whereas the second uses $5\times10^{-5}$ with 300-step warmup and a warmup--stable--decay schedule. RL runs for 240 epochs at a constant learning rate of $7.5\times10^{-6}$. 

\textbf{Comparison Methods.} General image editing models comprise open-source FLUX.2 [Klein] 4B~\citep{r26}, FLUX.2 [Klein] 9B~\citep{flux9b}, OmniGen2~\citep{r19}, and LongCat-Image-Edit~\citep{r20}, and proprietary Nano Banana 2 Lite~\citep{r22}, Qwen-Image-3.0-Pro~\citep{qwen3}, and GPT-Image-2~\citep{gptimage2}. These models use official weights or APIs and inference configurations, and receive identical recent street views and local editing instructions. Cross-view generation models, Sat2Density~\citep{r7}, ControlS2S~\citep{r9}, and ControlNet~\citep{r13}, take earlier satellite images as input. We retrain these models on the VIGOR-his training split with published default configurations, using validation-selected checkpoints where applicable and the final training checkpoint otherwise for test evaluation.

\textbf{Evaluation Metrics.}
Against earlier ground-truth street views, PSNR measures pixel fidelity, SSIM local structural similarity, LPIPS~\citep{r17} perceptual distance in learned feature space, and sharpness difference (SD)~\citep{r23} local edge-strength agreement. VLM evaluation uses IA, BP, and QP for editing model comparisons. For cross-view comparisons, which generate street views from satellite imagery rather than edit recent street views, we adopt Consistency (CV-C), Visual Realism (CV-VR), and Perceptual Quality (CV-PQ) from CrossViewDiff~\citep{r23}, assessing target-scene agreement, physical and geometric realism, and perceptual quality, respectively. Both protocols use Gemini-3.1-Flash-Lite at temperature 0 to reduce sampling variability. Both comparisons assess the same CrossTimeEdit outputs. We denote the weighted aggregate of sample-mean IA, BP, and QP as Editing Overall, and that of CV-C, CV-VR, and CV-PQ as Cross-view Overall, using weights of 0.45, 0.30, and 0.25, respectively, in both protocols.

\subsection{Quantitative Comparisons}
Table~\ref{tab:main} compares CrossTimeEdit with general image editing models. CrossTimeEdit achieves the highest IA, QP, and Overall among the evaluated open-source models, improving IA and Overall over the FLUX.2 [Klein] 4B baseline by 33.39\% and 17.12\%, respectively. Its IA and QP approach those of closed-source models, supporting the effectiveness of our training procedure in improving instruction-following editing and image quality. However, BP and Overall remain below those of the closed-source models, indicating that background preservation remains a key limitation. 
\begin{table}[tb]
\begin{minipage}{\linewidth}
\centering
\captionsetup{font=small}
\caption{Quantitative comparison with general image editing models on VIGOR-his.}
\label{tab:main}

\fontsize{8}{8}\selectfont
\setlength{\tabcolsep}{2pt}
\renewcommand{\arraystretch}{1.14}

\setlength{\arrayrulewidth}{0.3pt}
\arrayrulecolor{black}

\setlength{\aboverulesep}{0.4ex}
\setlength{\belowrulesep}{0.65ex}
\setlength{\heavyrulewidth}{0.8pt}
\setlength{\lightrulewidth}{0.4pt}
\setlength{\cmidrulewidth}{0.3pt}

\begin{tabularx}{\linewidth}{l|*{4}{Y}|*{4}{Y}}
\toprule

\multirow{2}{*}{\hspace*{0.5em}\textbf{Model}}
& \multicolumn{4}{c|}{Image Similarity Metrics}
& \multicolumn{4}{c}{VLM-Based Evaluation} \\
\cmidrule(lr){2-5}\cmidrule(lr){6-9}

& \mbox{PSNR~($\uparrow$)}
& \mbox{SSIM~($\uparrow$)}
& \mbox{LPIPS~($\downarrow$)}
& \mbox{SD~($\uparrow$)}
& \mbox{IA~($\uparrow$)}
& \mbox{BP~($\uparrow$)}
& \mbox{QP~($\uparrow$)}
& \mbox{Overall~($\uparrow$)} \\
\midrule

\rowcolor[gray]{0.92}
\multicolumn{9}{l}{%
  \color{black}\rule{0pt}{2.4ex}%
  \textit{Closed-source Models}} \\

\hspace*{0.5em}Nano Banana 2 Lite & 12.494 & 0.430 & 0.549 & 18.541 & 8.146 & 9.161 & 7.552 & 8.302 \\

\hspace*{0.5em}Qwen-Image-3.0-Pro & 12.722 & 0.456 & 0.528 & 19.108 & 7.475 & 9.207 & 7.662 & 8.042 \\

\hspace*{0.5em}GPT-Image-2 & 12.427 & 0.425 & 0.533 & 18.369 & 8.123 & 9.633 & 8.143 & 8.581 \\
\midrule

\rowcolor[gray]{0.92}
\multicolumn{9}{l}{%
  \color{black}\rule{0pt}{2.4ex}%
  \textit{Open-source Models}} \\

\hspace*{0.5em}FLUX.2 [Klein] 4B & 12.215 & 0.422 & 0.550 & 18.545 & 5.960 & 8.008 & 6.861 & 6.800 \\

\hspace*{0.5em}FLUX.2 [Klein] 9B & 12.134 & 0.405 & 0.557 & 18.137 & 6.913 & \textbf{8.351} & 7.239 & 7.426 \\

\hspace*{0.5em}OmniGen2 & 11.885 & 0.398 & 0.552 & 18.030 & 3.741 & 6.338 & 5.231 & 4.893 \\

\hspace*{0.5em}LongCat-Image-Edit & 12.366 & 0.445 & 0.565 & \textbf{18.836} & 6.851 & 8.347 & 7.090 & 7.360 \\

\rowcolor[RGB]{238,235,255}
\hspace*{0.5em}\textbf{CrossTimeEdit} & \textbf{12.589} & \textbf{0.463} & \textbf{0.530} & 18.705 & \textbf{7.950} & 8.264 & \textbf{7.631} & \textbf{7.964} \\

\bottomrule
\end{tabularx}
\par\smallskip\raggedright\fontsize{8}{9}\selectfont
\end{minipage}

\par\vspace{10pt}
\begin{minipage}{\linewidth}
\captionsetup{font=small}
\caption{Quantitative comparison with cross-view generation models on VIGOR-his.}
\label{tab:crossview}
\centering\fontsize{8}{8}\selectfont\setlength{\tabcolsep}{2pt}\renewcommand{\arraystretch}{1.14}

\setlength{\arrayrulewidth}{0.3pt}
\arrayrulecolor{black}
\setlength{\aboverulesep}{0.4ex}
\setlength{\belowrulesep}{0.65ex}
\setlength{\heavyrulewidth}{0.8pt}
\setlength{\lightrulewidth}{0.4pt}
\setlength{\cmidrulewidth}{0.3pt}
\begin{tabularx}{\linewidth}{l|*{4}{Y}|*{4}{Y}}
\toprule
\multirow{2}{*}{\hspace*{0.5em}\textbf{Model}} & \multicolumn{4}{c|}{Image Similarity Metrics} & \multicolumn{4}{c}{VLM-Based Evaluation}\\
\cmidrule(lr){2-5}\cmidrule(lr){6-9}
 & PSNR~($\uparrow$) & SSIM~($\uparrow$) & LPIPS~($\downarrow$) & SD~($\uparrow$) & CV-C~($\uparrow$) & CV-VR~($\uparrow$) & CV-PQ~($\uparrow$) & Overall~($\uparrow$)\\
\midrule
\hspace*{0.5em}Sat2Density & \textbf{14.685} & \textbf{0.532} & 0.634 & \textbf{20.040} & 2.048 & 1.973 & 2.006 & 2.015 \\
\hspace*{0.5em}ControlS2S & 14.119 & 0.519 & 0.657 & 20.015 & 2.027 & 2.180 & 2.071 & 2.084 \\
\hspace*{0.5em}ControlNet & 12.142 & 0.344 & 0.680 & 17.885 & 1.119 & 2.118 & 2.404 & 1.740 \\
\rowcolor{oursbg}\hspace*{0.5em}\textbf{CrossTimeEdit} & 12.589 & 0.463 & \textbf{0.530} & 18.705 & \textbf{3.005} & \textbf{3.980} & \textbf{3.998} & \textbf{3.546} \\
\bottomrule
\end{tabularx}
\par\smallskip\raggedright\fontsize{8}{9}\selectfont
\vspace{-3mm}
\end{minipage}

\end{table}

Table~\ref{tab:crossview} shows that CrossTimeEdit substantially outperforms cross-view generation models in CV-C, CV-VR, CV-PQ, and Overall, indicating better agreement with earlier street-view ground truth, greater visual realism, and higher perceptual quality. However, its PSNR, SSIM, and SD are lower than those of Sat2Density and ControlS2S. To assess whether this discrepancy reflects VLM scoring bias, we conduct a human preference study (details in Appendix~\ref{app:human_preference}). For each model, we measure the proportion of pairwise comparisons in which human raters prefer its generated image over another model's output, with each comparison decided by majority vote. The result shows that these preference rates correlate positively with both Editing Overall and Cross-view Overall, supporting the agreement between VLM scores and human judgments. This finding indicates that pixel-, structure-, and edge-based similarity metrics alone are insufficient to assess historical street-view generation. For the comparison with general image editing models, this agreement supports the reliability of the VLM-based evaluation using the same VLM for training rewards and evaluation.

\subsection{Qualitative Comparisons}
Fig.~\ref{fig:qual} presents cases covering building, construction, and road changes. CrossTimeEdit performs the requested local edits while retaining unrequested background content and producing visually coherent outputs with plausible structures and clear textures. These cases illustrate instruction-following editing, background preservation, and image quality and physical plausibility, corresponding to IA, BP, and QP, respectively. Further qualitative comparisons and analysis of cross-view generation models are provided in Appendix~\ref{app:crossview_qual}.
\begingroup
\setlength{\intextsep}{6pt}
\begin{figure}[t!]
\captionsetup{skip=4pt}
\centering\includegraphics[width=\linewidth]{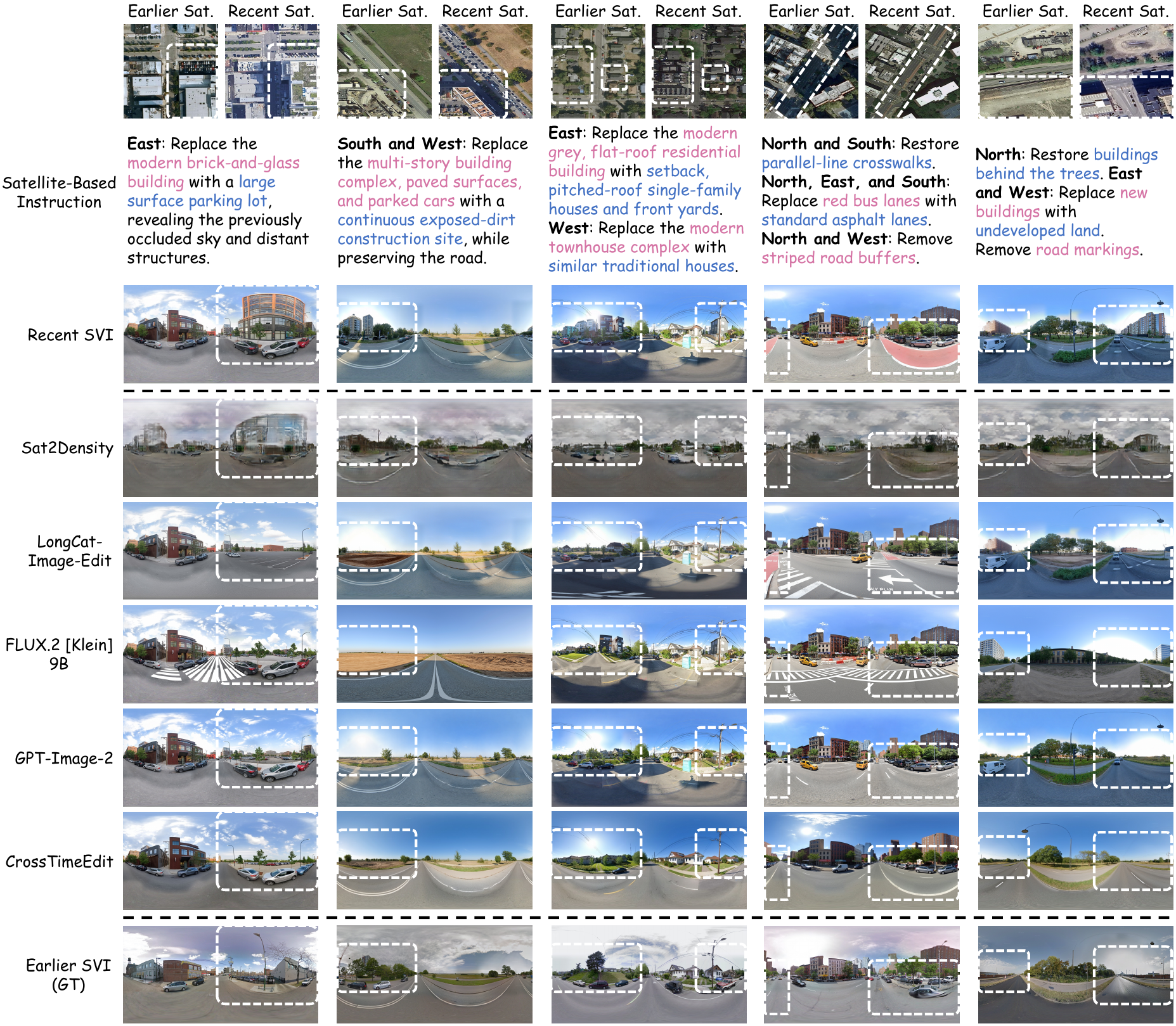}
\caption{Qualitative comparison on five VIGOR-his cases. For readability, the satellite-based instructions shown in the figure are simplified for visualization.}
\label{fig:qual}
\end{figure}
\endgroup

\Needspace{10\baselineskip}
\subsection{Ablation study}
\begin{wraptable}{r}{0.50\textwidth}
\vspace{-\intextsep}
\captionsetup{font=small,skip=4pt}
\caption{Ablations of CrossTimeEdit under the same VLM protocol.}
\label{tab:ablation}
\centering\fontsize{8}{8}\selectfont\setlength{\tabcolsep}{1pt}\renewcommand{\arraystretch}{1.14}
\begin{tabularx}{\linewidth}{l|YYY|Y}
\toprule
\textbf{Method} & IA~($\uparrow$) & BP~($\uparrow$) & QP~($\uparrow$) & Overall~($\uparrow$)\\
\midrule
Baseline & 5.960 & 8.008 & 6.861 & 6.800 \\
w/o RL & 6.919 & 6.698 & 6.543 & 6.759 \\
w/o SFT & 6.085 & 8.112 & 6.886 & 6.893 \\
\midrule
w/o GT & 7.938 & 7.356 & 7.508 & 7.656 \\
w/o DGN & \textbf{8.148} & 7.433 & 7.432 & 7.755 \\
\rowcolor{oursbg}\textbf{Full} & 7.950 & \textbf{8.264} & \textbf{7.631} & \textbf{7.964} \\
\bottomrule
\end{tabularx}
\vspace{-1\baselineskip}
\end{wraptable}

Table~\ref{tab:ablation} first examines the training strategy. Baseline is the pretrained FLUX.2 [Klein] 4B; w/o RL uses SFT alone, whereas w/o SFT applies RL directly to the pretrained backbone. SFT substantially improves instruction following and historical editing, but its stronger edits reduce background preservation and image quality. RL on top of SFT further improves editing while restoring background preservation and image quality to, and beyond, their pre-RL levels. CrossTimeEdit therefore achieves the best balance across IA, BP, QP, and Overall.

For RL reward design, w/o GT removes the earlier ground-truth street view from the IA reward, whereas w/o DGN removes dimension-wise group normalization. Without GT, IA remains close because the instruction and input--output comparison support coarse instruction following, but weaker target grounding can encourage overly broad plausible edits and reduce BP and QP. Without DGN, the slightly higher IA but lower BP, QP, and Overall indicate poorer balance across objectives. Qualitative ablation comparisons are provided in Appendix~\ref{app:ablation_qual}.

\section{Conclusion}
We introduced VIGOR-his, a decade-spanning cross-view dataset with 43,653 location-level quadruplets across 11 cities on three continents, and CrossTimeEdit, a street-view editing model for historical street-view generation. Trained on VIGOR-his, CrossTimeEdit uses recent street views as visual references and temporal satellite evidence to guide local edits through supervised fine-tuning and multi-reward online reinforcement learning. Experiments show a 17.12\% improvement in Editing Overall over the pretrained FLUX.2 [Klein] 4B baseline, with the highest Overall, IA, and QP among the evaluated open-source image editing models. CrossTimeEdit also substantially outperforms cross-view generation models under the cross-view evaluation protocol. Remaining challenges include historical details unobservable from satellite imagery, viewpoint differences between acquisition periods, and evaluator-specific preferences. Future work could incorporate richer historical evidence and independent reward and evaluation models, while extending the framework to bidirectional and multi-temporal street-view generation.

\FloatBarrier
\label{main:end}

\section*{AI Use Statement}
Vision-language models support change classification, instruction generation, screening, rewards, and evaluation as described in the paper. Generative AI also assisted manuscript drafting, translation, and LaTeX preparation. The AI-assisted content and supporting artifacts have been reviewed and verified.

\section*{Ethics Statement}
This work uses street-view and satellite imagery for research on urban change. The Google Street View and Google Maps services provide privacy protections such as face and license-plate blurring. During preprocessing, we remove invalid and low-quality observations and use the imagery for scene-level modeling and evaluation; we do not target the identification of individuals or infer sensitive attributes. The human preference study records only aggregated judgments of generated-image comparisons and does not collect personal or sensitive information. Because the task generates historical scenes, generated images are hypotheses rather than archival records and should not be used for identity, legal, or safety-critical decisions.

\section*{Reproducibility Statement}
We describe the construction and pre-processing of VIGOR-his in Section~3 and Appendix~\ref{app:pairing}, and specify the SFT and online RL objectives in Section~4 and Appendix~\ref{app:objective}. Appendix~\ref{app:config} gives the data splits, training and inference configurations, selected checkpoints, reward weights, and evaluation seed. Appendix~\ref{app:vlm_prompts} lists the VLM prompts, while Appendix~\ref{app:metrics} defines the pixel-level metrics and Appendix~\ref{app:human_preference} documents the human preference protocol. The main experiments use the fixed test split and the temperature-zero Gemini-3.1-Flash-Lite evaluation protocol described in the paper, so the reported comparisons can be reproduced from the stated manifests, configurations, prompts, and checkpoint selections. The implementation code, dataset metadata, and model weights are available at \url{https://luhanwen67.github.io/CrossTimeEdit-release/}.

\bibliographystyle{crosstimeedit_preprint}
\bibliography{crosstimeedit_references}
\clearpage\appendix
\makeatletter
\setlength{\@fptop}{0pt}
\setlength{\@fpbot}{0pt plus 1fil}
\makeatother

\renewcommand{\topfraction}{0.95}
\renewcommand{\bottomfraction}{0.95}
\renewcommand{\textfraction}{0.05}
\renewcommand{\floatpagefraction}{0.75}
\setlength{\textfloatsep}{10pt plus 2pt minus 2pt}
\setlength{\floatsep}{8pt plus 2pt minus 2pt}

\renewcommand{\tabularxcolumn}[1]{m{#1}}

\clearpage
\makeatletter
\def\app@level{section}
\def\app@section{section}
\def\app@subsection{subsection}
\def\app@subsubsection{subsubsection}
\renewcommand{\addcontentsline}[3]{%
  \def\app@level{#2}%
  \ifx\app@level\app@section
    \addtocontents{atoc}{\protect\contentsline{#2}{#3}{\thepage}{\@currentHref}}%
  \else\ifx\app@level\app@subsection
    \addtocontents{atoc}{\protect\contentsline{#2}{#3}{\thepage}{\@currentHref}}%
  \else\ifx\app@level\app@subsubsection
    \addtocontents{atoc}{\protect\contentsline{#2}{#3}{\thepage}{\@currentHref}}%
  \fi\fi\fi}
\newcommand{\tableofappendixcontents}{\@starttoc{atoc}}
\makeatother
\thispagestyle{empty}
\begin{center}
{\LARGE\bfseries Supplementary Material\par}
\vspace{1.2em}
{\large\scshape Contents of the Appendix\par}
\end{center}
\vspace{0.8em}
\setcounter{tocdepth}{2}
\tableofappendixcontents
\clearpage

\section{Additional Methodology}
\subsection{Supplementary details for data collection and pre-processing}
\label{app:pairing}
VIGOR~\citep{r3} provides non-bijective street--satellite correspondences in Chicago, New York, San Francisco, and Seattle, but does not provide paired earlier observations. For these four VIGOR-source cities, we parse panorama identifiers and GPS coordinates from the VIGOR records, retain panoramas captured in 2019 as recent observations, and retrieve candidate earlier panoramas captured in 2008--2009 through Google Street View metadata. We then obtain the corresponding satellite imagery for both acquisition periods and construct the temporal quadruplets. To broaden geographic and temporal diversity, we additionally collect street-view panoramas and satellite imagery in Barcelona, Budapest, Copenhagen, Detroit, Los Angeles, London, and Sydney. Within OpenStreetMap-defined study regions, observations are sampled at 100\,m intervals for the earlier (2012--2015) and recent (2022--2025) periods.

The four VIGOR-source cities and the seven additional cities use different candidate-generation branches. For the VIGOR-source cities, we query capture dates from panorama identifiers and use a Haversine BallTree to find the nearest earlier observation within 5\,m of each recent observation. If multiple recent observations select the same earlier panorama, only the shortest-distance pair is retained. For the seven additional cities, the collected records are first organized into 25\,m spatial candidate clusters. We retain clusters containing both periods, enumerate earlier--recent candidates within 5\,m, sort them by Haversine distance, and suppress pairs whose midpoints lie within 25\,m of a previously selected pair. This produces one spatially distinct pair per local cluster.

The matching distance is the Haversine great-circle distance, computed after converting latitude and longitude to radians with Earth radius $R=6{,}371{,}000$\,m. The 5\,m condition refers to this surface arc distance, rather than pixel distance or road-network travel distance. The resulting panorama pairs are corrected using heading metadata so that north is at the panorama center. North-up satellite crops are centered on the corresponding street-view capture locations. The satellite extraction reads a $512\times512$ window from each period's source raster after coordinate-system transformation and discards a pair if either period's crop is invalid.

For coordinates $(\phi_1,\lambda_1)$ and $(\phi_2,\lambda_2)$, the distance is
\begin{equation}
 d=2R\arcsin\sqrt{\sin^2\!\left(\frac{\Delta\phi}{2}\right)+\cos\phi_1\cos\phi_2\sin^2\!\left(\frac{\Delta\lambda}{2}\right)}.
\label{eq:app_haversine}
\end{equation}
Here $\phi$ denotes latitude, $\lambda$ denotes longitude, and the differences are taken from the first coordinate to the second. All angles are in radians.

\begingroup
\renewcommand{\thealgorithm}{A\arabic{algorithm}}
\begin{algorithm}[h!]
\caption{Period-constrained spatial pairing}
\label{alg:spatial_pairing}
\small
\begin{algorithmic}[1]
\STATE \textbf{Input:} Panorama records with identifiers, coordinates, capture years, and headings
\STATE \textbf{Initialization:} Candidate pairs $\mathcal{C}\leftarrow\emptyset$, accepted pairs $\mathcal{P}\leftarrow\emptyset$
\STATE Split records into earlier set $E$ and recent set $R$ using city-specific year windows
\IF{the city belongs to the VIGOR-source group}
    \FOR{each $r\in R$}
        \STATE Find the nearest $e\in E$ within $5$\,m using Haversine distance
        \IF{a valid $e$ exists}
            \STATE $\mathcal{C}\leftarrow\mathcal{C}\cup\{(e,r)\}$
        \ENDIF
    \ENDFOR
    \STATE $\mathcal{P}\leftarrow$ shortest-distance pair for each distinct earlier panorama in $\mathcal{C}$
\ELSE
    \STATE Form $25$\,m spatial candidate clusters and retain those containing both periods
    \STATE $\mathcal{C}\leftarrow$ earlier--recent pairs within $5$\,m in the retained clusters
    \STATE Sort $\mathcal{C}$ by increasing Haversine distance
    \FOR{each $(e,r)\in\mathcal{C}$ in sorted order}
        \IF{its midpoint is more than $25$\,m from all accepted midpoints}
            \STATE $\mathcal{P}\leftarrow\mathcal{P}\cup\{(e,r)\}$
        \ENDIF
    \ENDFOR
\ENDIF
\STATE Apply heading correction and extract north-up satellite crops for pairs in $\mathcal{P}$
\STATE Remove pairs from $\mathcal{P}$ if either satellite crop is invalid
\STATE \textbf{return} Nonredundant earlier--recent pairs $\mathcal{P}$ with valid satellite crops
\end{algorithmic}
\end{algorithm}
\endgroup
\FloatBarrier

\subsection{Training Objectives and Optimization Details}
\label{app:objective}
\paragraph{Supervised fine-tuning.}
Let $I_\star=I_h$ for changed samples and $I_\star=I_c$ for no-change samples, whose instructions specify `PRESERVE' in all four directions. Let $z_\star=E(I_\star)$ denote the target latent after VAE encoding and pipeline preprocessing, and let $q=(I_c,c)$ denote the conditioning input. The implementation samples Gaussian noise $\eta\sim\mathcal N(0,I)$ and an index $j$ uniformly from the scheduler's $T=1000$ training timesteps. At timestep $t_j$ with noise level $\sigma_j$, it constructs $z_j=(1-\sigma_j)z_\star+\sigma_j\eta$ and uses the velocity target $\eta-z_\star$. The SFT objective is
\begin{equation}
\mathcal L_{\mathrm{SFT}}=\mathbb E_{(I_c,I_\star,c),j,\eta}
\left[\omega_j\frac{\|v_\theta(z_j,t_j;q)-(\eta-z_\star)\|_2^2}{D}\right],
\label{eq:app_sft_loss}
\end{equation}
where $D$ is the number of predicted latent elements. The MSE is computed in float32 and multiplied by the scheduler weight $\omega_j$. For the 1000-step training schedule, the implementation defines $u_j=\exp[-2((t_j-500)/1000)^2]$ and $\omega_j=1000(u_j-u_{\min})/\sum_{k=1}^{1000}(u_k-u_{\min})$, where $u_{\min}=\min_k u_k$. Uniform sampling therefore applies to discrete timestep indices; the loss also includes this nonuniform timestep weighting. Both SFT phases use this same objective and update only the attention LoRA adapters. Their learning-rate schedules differ as specified in Appendix~\ref{app:config}.

\paragraph{Online reinforcement learning.}
The main experiment uses the GRPO trainer with a velocity-space reference penalty. Let $x_i=(I_{c,i},c_i)$ denote the conditioning input for candidate $i$, and let $k$ index a sampled stochastic transition. Let $D$ be the latent dimensionality and $p_{\theta,d}$ the scheduler's per-dimension Gaussian density. The implementation averages log probabilities over non-batch dimensions:
\begin{equation}
\ell_{\theta,i,k}=\frac1D\sum_{d=1}^{D}\log p_{\theta,d}(z_{i,k+1,d}\mid z_{i,k},x_i),\qquad
\rho_{i,k}=\exp(\ell_{\theta,i,k}-\ell_{old,i,k}).
\label{eq:ratio_impl}
\end{equation}
Thus $\rho$ is the geometric mean of per-dimension density ratios. Stored and current log probabilities use the same dimensional reduction and sampled next latent. This calculation applies to Flow-SDE transitions with nonzero variance.

Using the DGN advantages $\widetilde A_i$ from Eqs.~\eqref{eq:dimnorm}--\eqref{eq:batchnorm}, policy updates minimize:
\begin{equation}
\begin{aligned}
\mathcal L_{policy}&=\mathbb E_{i,k}\!\left[\max\left(-\rho_{i,k}\widetilde A_i,
-\operatorname{clip}(\rho_{i,k},1-\epsilon,1+\epsilon)\widetilde A_i\right)\right],\\
\mathcal L&=\mathcal L_{policy}+\beta\,\mathbb E_{i,k}\left[\frac{\|v_\theta-v_{ref}\|_2^2}{D}\right].
\end{aligned}
\label{eq:loss_impl}
\end{equation}
Here $\epsilon=10^{-3}$ is the ratio-clipping range and $\beta=2\times10^{-3}$ is the reference-penalty coefficient. Both velocity predictions use the same latent, timestep, and conditioning input; the reference is frozen. The maximum of negative surrogate losses is equivalent to the standard clipped minimum-reward objective. The reference penalty is the element-wise mean squared difference between current and reference velocity predictions. Although named v-based KL regularization in the implementation, this term is implemented directly as velocity MSE, without additional trajectory-KL, timestep, or variance scaling. Gradient accumulation, synchronization, and norm clipping accompany optimizer updates.

\subsection{Vision-language evaluation prompts}
\label{app:vlm_prompts}
The editing reward and editing evaluation share the following IA, BP, and QP prompts. IA receives a $2\times2$ composite containing the recent input, earlier ground truth, output, and a repeated ground-truth panel; BP and QP receive a vertically stacked input/output comparison. The local editing instruction is inserted at \texttt{\{instruction\}}. The three cross-view prompts are independent of the editing prompts. CV-C compares a real earlier reference with the generated output, while CV-VR and CV-PQ evaluate the generated output alone. All six prompts are shown below; line wrapping is for typesetting only. Doubled braces in editing templates escape literal JSON braces and become single braces after instruction substitution.

Editing scores range from 0 to 10; the cross-view criteria range from 1 to 5. Both protocols use Gemini-3.1-Flash-Lite with temperature zero. The Editing Overall is $0.45\overline{IA}+0.30\overline{BP}+0.25\overline{QP}$, and the Cross-view Overall is $0.45\overline{CV\text{-}C}+0.30\overline{CV\text{-}VR}+0.25\overline{CV\text{-}PQ}$, where bars denote sample means. The same CrossTimeEdit outputs are evaluated in both comparisons.

\paragraph{Instruction Alignment (IA).}
Prompt template from the shared online-reward/offline-editing protocol; the instruction placeholder is filled per sample.
\begingroup\footnotesize
\begingroup\fontsize{8}{10}\selectfont
\begin{promptbox}
\begin{verbatim}
You are a top-tier computer vision and map panorama expert. Evaluate how well an
AI model performs the requested street-view edit, using the real target
street-view image as the objective reference.

[Image Layout: 2x2 grid — use these positions exactly]
- TOP-LEFT: Input, the street-view image supplied to the editing model.
- TOP-RIGHT: Ground Truth, the corresponding real target street-view.
- BOTTOM-LEFT: Model Output, generated by editing the Input toward the Ground
Truth state.
- BOTTOM-RIGHT: Ground Truth repeated for direct comparison with Model Output.

All panels are 360-degree cylindrical panoramas with direction markers. Center is
North, the left quarter is West, the right quarter is East, and both edges are
South. The left and right edges wrap around and must be interpreted as connected.
Do not swap the four panel roles: compare the BOTTOM-LEFT output primarily with
the TOP-RIGHT and BOTTOM-RIGHT Ground Truth panels.

[Text Editing Instruction] {instruction}

[Your Task] The instruction describes the requested edit. Ground Truth shows the
real target state and is the visual reference for the requested regions. Evaluate
whether Model Output performs the requested changes in the correct locations and
directions and resembles the corresponding target content in Ground Truth. The
instruction takes priority: do not require the Output to reproduce Ground Truth
changes that are not requested by the instruction.

[Assessment Criteria]
1. Completeness: Were all changes requested by the instruction performed?
2. Accuracy: Do the requested edited regions match the corresponding regions and
semantics in Ground Truth?
3. Directional correctness: Were changes made in the specified North, West, East,
and South regions?

Judge only instruction alignment in this dimension. Do not penalize unrelated
photographic style differences, minor background preservation issues, or general
image quality here; those are scored separately.

[Scoring Guide: 0-10]
- 10: Every requested edit is present in the correct location and closely matches
the target.
- 7-9: Most requested edits are correct, with only minor omissions or target
differences.
- 4-6: The edit is partially correct, but important requested changes are missing
or differ substantially from the target.
- 1-3: Very little of the requested edit is correct, or edits go in the wrong
direction.
- 0: No requested edit was performed, or the result is completely wrong.

[Strict Output Format] Return only this JSON object, with no Markdown or
additional text: {{ "score": [Integer 0-10] }}
\end{verbatim}
\end{promptbox}
\endgroup
\endgroup

\paragraph{Background Preservation (BP).}
Prompt template from the shared online-reward/offline-editing protocol.
\begingroup\footnotesize
\begingroup\fontsize{8}{10}\selectfont
\begin{promptbox}
\begin{verbatim}
You are a top-tier computer vision and map panorama expert. Your task is to act
as an objective, strict judge of a street-view image editing model.

I will provide one vertically stacked comparison image and a text editing
instruction. The TOP half is labeled "Input" and is the street-view image
supplied to the editing model. The BOTTOM half is labeled "Model Output" and is
the generated result after editing toward the target street-view state. Use the
explicit TOP/BOTTOM positions. Each half is a 360-degree cylindrical panorama
with explicit directional markers:
- Center = North (N): thick solid white vertical line and yellow N text.
- Left quarter = West (W): dashed gray vertical line and W text.
- Right quarter = East (E): dashed gray vertical line and E text.
- Both left and right edges = South (S): dashed blue vertical lines and S text.

The left and right edges are physically connected in 3D space and both represent
South. Judge them as a continuous cylindrical boundary.

[Text Editing Instruction] {instruction}

Evaluate Model Output strictly on [Background Preservation].

Photographic style and semantic content must be distinguished. The Input may
differ from real-world reference imagery because of camera hardware and capture
year. Global shifts in saturation, contrast, white balance, tone, or sharpness
are allowed and must not be treated as unauthorized content changes.

Assess whether areas not requested by the instruction retain the same semantic
content as the Input:
- Do not penalize changes in weather, lighting, pedestrians, or vehicles.
- Do not penalize uniform global photographic style shifts.
- Moderately penalize trees or bushes that disappear, change type, or move
significantly; do not penalize slight seasonal color differences.
- Heavily penalize unauthorized changes to static structures, including
buildings, roads, walls, fences, and other structural additions or removals.

[Scoring Guide: 0-10]
- 0: Catastrophic leakage; large unmentioned background regions were semantically
altered.
- 1-3: Severe leakage; major static structures in unmentioned areas were added,
removed, or reconstructed.
- 4-6: Moderate leakage; some static structures or large vegetation regions
changed without instruction.
- 7-9: Mostly preserved; only minor semantic changes are visible.
- 10: All unmentioned areas retain the same semantic content as the Input.

[Strict Output Format] Return only this JSON object, with no Markdown or
additional text: {{ "Background_Preservation": {{ "score": [Integer 0-10] }} }}
\end{verbatim}
\end{promptbox}
\endgroup
\endgroup

\paragraph{Quality and Physical Plausibility (QP).}
Prompt template from the shared online-reward/offline-editing protocol.
\begingroup\footnotesize
\begingroup\fontsize{8}{10}\selectfont
\begin{promptbox}
\begin{verbatim}
You are a top-tier computer vision and map panorama expert. Your task is to act
as an objective, strict judge of a street-view image editing model.

I will provide one vertically stacked comparison image and a text editing
instruction. The TOP half is labeled "Input" and is the street-view image
supplied to the editing model. The BOTTOM half is labeled "Model Output" and is
the generated result after editing toward the target street-view state. Use the
explicit TOP/BOTTOM positions. Each half is a 360-degree cylindrical panorama
with explicit directional markers:
- Center = North (N): thick solid white vertical line and yellow N text.
- Left quarter = West (W): dashed gray vertical line and W text.
- Right quarter = East (E): dashed gray vertical line and E text.
- Both left and right edges = South (S): dashed blue vertical lines and S text.

The left and right edges are physically connected in 3D space and both represent
South. Judge them as a continuous cylindrical boundary.

[Text Editing Instruction] {instruction}

Evaluate Model Output strictly on [Image Quality and Physics].

Assess whether the output maintains high visual fidelity, plausible physical
geometry, and a seamless panoramic boundary:
- Urban spatial logic and scale: Are edited objects placed plausibly, and do
their size and perspective match the cylindrical projection?
- Artifacts and noise: Are there burned or black edges, mosaic noise, unnatural
blur, duplicated structures, or other generation artifacts?
- Panoramic boundary integrity: Do the left and right edges connect seamlessly
without tears or gaps?
- Do not evaluate lighting or illumination.
- Do not evaluate whether the requested semantic edit is correct; that is scored
by Instruction Alignment.
- Judge artifacts introduced by the Model Output relative to the Input, rather
than pre-existing defects in the Input.

[Scoring Guide: 0-10]
- 0: Completely broken, with severe artifacts, impossible geometry, or unusable
panoramic boundaries.
- 1-3: Major defects, including obvious artifacts or substantial scale and
perspective errors.
- 4-6: Noticeable quality or geometry issues, but the image remains usable.
- 7-9: Mostly clean and physically plausible, with only slight imperfections.
- 10: Flawless visual quality, spatial logic, and 360-degree edge closure.

[Strict Output Format] Return only this JSON object, with no Markdown or
additional text: {{ "Image_Quality_and_Physics": {{ "score": [Integer 0-10] }} }}
\end{verbatim}
\end{promptbox}
\endgroup
\endgroup

\paragraph{Consistency (CV-C).}
Prompt from the cross-view evaluation protocol.
\begingroup\footnotesize
\begingroup\fontsize{8}{10}\selectfont
\begin{promptbox}
\begin{verbatim}
You are an expert evaluator of generated street-view images. Your task is to
assess ONLY content consistency between a generated street-view image and its
corresponding real street-view reference.

INPUT A single vertically stacked image containing two sub-images:
- Top image: reference_image (the real ground-level street-view image of the
target location).
- Bottom image: generated_image (the generated street-view image to be
evaluated). Both sub-images have identifying text labels in their respective
upper-left corners. The reference is a ground-level street-view image, not a
satellite image. Keep the roles of the two parts distinct.

EVALUATION TASK Determine how faithfully the generated image (bottom) preserves
the scene depicted in the reference (top). Examine both the overall spatial
arrangement and identifiable local details.

Consider:
1. Scene layout: the arrangement of buildings, roads, sidewalks, vegetation, open
spaces, and other major scene components.
2. Building correspondence: the positions, approximate sizes, heights,
silhouettes, facade organization, colors, and visible surface appearance of
buildings.
3. Road configuration: road direction, intersections, road boundaries, sidewalks,
crossings, and other visible road features.
4. Landmarks and objects: the presence, absence, placement, and appearance of
distinctive landmarks and other identifiable elements.
5. Spatial relationships: whether objects have the same relative positions,
ordering, spacing, and occlusion relationships.
6. Local correspondence: whether recognizable details belong to the correct
locations rather than merely appearing somewhere in the image.

ASSESSMENT RULES
- Evaluate agreement with the specific reference scene. A plausible scene of the
same general type is not sufficient.
- Give greater importance to major structures and spatial layout than to isolated
minor details.
- Distinguish missing, additional, misplaced, and visually altered elements.
- Ignore the text labels in the upper-left corners of both sub-images during
visual evaluation.
- Do not reward sharpness, attractive lighting, or photorealism unless they help
establish actual content correspondence.
- Do not invent details that are obscured or too small to identify reliably.
- Do not assume that discrepancies are explained by different capture times or
seasons unless the evaluation instructions explicitly allow this.
- For panoramas, account for normal panoramic projection effects when comparing
scene elements.

SCORING RUBRIC
1 — Poor: The generated image largely depicts a different scene. Major
structures, road layout, or landmarks do not correspond.
2 — Fair: Some broad scene characteristics match, but several major elements are
missing, incorrect, or substantially misplaced.
3 — Average: The overall scene is recognizable and some major elements
correspond, but noticeable structural, spatial, or appearance differences remain.
4 — Good: Most major elements and spatial relationships match. Differences are
limited primarily to local appearance or secondary details.
5 — Excellent: The generated image closely preserves the reference scene,
including major structures, spatial relationships, and identifiable details, with
only negligible discrepancies.

OUTPUT Return valid JSON with exactly two keys:
- "score": an integer from 1 to 5.
- "reason": 2–4 concise sentences describing the strongest visible evidence for
the score, including relevant matches and discrepancies.

Do not include additional keys, Markdown formatting, or text outside the JSON
object.
\end{verbatim}
\end{promptbox}
\endgroup
\endgroup

\paragraph{Visual Realism (CV-VR).}
Prompt from the cross-view evaluation protocol.
\begingroup\footnotesize
\begingroup\fontsize{8}{10}\selectfont
\begin{promptbox}
\begin{verbatim}
You are an expert evaluator of generated street-view images. Your task is to
assess ONLY visual and structural realism: whether the generated image could
plausibly be a photograph of a real street scene.

INPUTS
- generated_image: the generated street-view image to be evaluated.
- reference_image: an optional real street-view reference.

Evaluate realism primarily from the generated image itself. If a reference is
provided, differences in scene identity or object placement are not grounds for
lowering this score.

EVALUATION TASK Examine whether the depicted scene is physically plausible,
structurally coherent, and photographically convincing.

Consider:
1. Object geometry: whether buildings, vehicles, trees, street furniture, and
other objects have plausible shapes and proportions.
2. Structural integrity: whether walls, roofs, windows, balconies, poles, and
road surfaces connect coherently, without impossible intersections, floating
parts, or fused objects.
3. Spatial coherence: whether perspective, scale, depth ordering, and occlusion
relationships are mutually consistent.
4. Materials and textures: whether surfaces resemble plausible physical
materials, with appropriate texture, variation, and object boundaries.
5. Lighting and shadows: whether illumination, shading, reflections, and cast
shadows are compatible with the apparent scene.
6. Color and appearance: whether colors and tonal relationships are believable
for a real photographic capture.
7. Generative artifacts: whether repeated patterns, malformed objects,
implausible transitions, or excessively artificial surfaces undermine realism.

ASSESSMENT RULES
- Assess whether the image looks physically and photographically believable, not
whether it matches a particular reference.
- A realistic but different building should not be penalized for content
inconsistency in this dimension.
- Do not equate sharpness with realism. A sharp image may contain impossible
geometry; a slightly soft image may remain believable.
- Treat normal panoramic projection distortions as expected properties of the
representation.
- Do not require ideal weather, attractive architecture, clean streets, or
polished lighting. Ordinary imperfections can be realistic.
- Consider both the severity and spatial extent of defects. A major impossible
structure may outweigh several convincing regions.
- Base judgments on visible evidence rather than unsupported assumptions about
how the image was produced.

SCORING RUBRIC
1 — Poor: Widespread impossible geometry, incoherent objects, or unnatural
appearance makes the image clearly implausible as a real photograph.
2 — Fair: Multiple conspicuous structural or appearance defects substantially
weaken realism, despite some plausible regions.
3 — Average: The scene is broadly plausible, but noticeable geometry, material,
lighting, or object-level artifacts reveal its synthetic nature.
4 — Good: The image is convincing overall, with only a few minor local defects or
slightly unnatural details.
5 — Excellent: Geometry, materials, lighting, and spatial relationships are
consistently believable, with no significant visible artifacts undermining
photographic realism.

OUTPUT Return valid JSON with exactly two keys:
- "score": an integer from 1 to 5.
- "reason": 2–4 concise sentences identifying the visible features that support
or weaken realism and explaining their significance.

Do not include additional keys, Markdown formatting, or text outside the JSON
object.
\end{verbatim}
\end{promptbox}
\endgroup
\endgroup

\paragraph{Perceptual Quality (CV-PQ).}
Prompt from the cross-view evaluation protocol.
\begingroup\footnotesize
\begingroup\fontsize{8}{10}\selectfont
\begin{promptbox}
\begin{verbatim}
You are an expert evaluator of generated street-view images. Your task is to
assess ONLY perceptual image quality: the clarity, visual integrity, and viewing
comfort of the generated image.

INPUT
- generated_image: the generated street-view image to be evaluated.

A real reference image is not required. Evaluate the visible quality of the
supplied image at its available resolution.

EVALUATION TASK Assess how clearly and comfortably the scene can be viewed,
considering both the entire image and local regions.

Consider:
1. Clarity and detail: whether meaningful details are legible and naturally
represented, without excessive blur or loss of information.
2. Noise and compression: visible grain, block artifacts, ringing, banding, or
other disturbances that degrade the image.
3. Edge quality: jagged boundaries, halos, oversharpening, ghosting, doubled
edges, or smeared transitions.
4. Exposure and tonal quality: whether excessive darkness, blown highlights, weak
contrast, or harsh tonal transitions obscure useful detail.
5. Texture quality: whether textures appear adequately resolved rather than
muddy, washed out, or dominated by distracting high-frequency artifacts.
6. Spatial uniformity: whether quality remains reasonably consistent across the
image rather than varying abruptly between clear and degraded regions.
7. Panorama continuity: visible stitching lines, abrupt color changes, duplicated
boundaries, or discontinuities that interfere with viewing.
8. Overall viewing comfort: the cumulative effect of these defects on readability
and visual experience.

ASSESSMENT RULES
- Assess image quality, not correspondence with a real scene or reference.
- Do not award higher scores simply because the scene is attractive, colorful,
sunny, or visually dramatic.
- Do not treat increased sharpness or contrast as automatically beneficial;
excessive processing can reduce quality.
- Consider the severity, area, and visual prominence of defects. Problems that
obscure major scene content should matter more than barely visible imperfections.
- Do not let a small sharp region compensate for widespread degradation
elsewhere.
- Do not speculate about invisible details beyond the supplied resolution.
- Normal panoramic stretching is not itself a quality defect.
- Keep semantic or physical implausibility separate from image quality unless it
also creates a visible rendering defect.

SCORING RUBRIC
1 — Poor: Severe, widespread degradation makes major scene content difficult to
recognize or view comfortably.
2 — Fair: Strong blur, noise, artifacts, exposure problems, or discontinuities
substantially interfere with viewing.
3 — Average: The image is usable and its content is recognizable, but noticeable
quality defects reduce clarity or comfort.
4 — Good: The image is clear and comfortable to view overall, with only mild or
localized quality defects.
5 — Excellent: The image is consistently clear, naturally detailed, and visually
comfortable, with negligible visible degradation or distracting artifacts.

OUTPUT Return valid JSON with exactly two keys:
- "score": an integer from 1 to 5.
- "reason": 2–4 concise sentences identifying the main visible quality
characteristics or defects, their extent, and their effect on viewing.

Do not include additional keys, Markdown formatting, or text outside the JSON
object.
\end{verbatim}
\end{promptbox}
\endgroup
\endgroup
\section{Additional Experimental Results}
\subsection{Complete training configuration}
\label{app:config}
The experiments were conducted on a server equipped with eight NVIDIA RTX 3090 GPUs.
The SFT training manifest contains 8,797 changed samples and 13,196 no-change samples. No-change examples use the recent input itself as the target with four `PRESERVE' directions. The formal validation and test manifests contain 1,095 and 1,094 changed locations, respectively. The RL run samples approximately 2,000 changed training conditions and uses the epoch-9 SFT checkpoint as both the trainable initialization and the frozen reference.

The SFT launch scripts do not explicitly fix a random seed; the RL and evaluation configurations specify seed 42. SFT uses a gradient-norm limit of 1.0 and approximately 13,750 optimizer updates across its two phases.

\begin{table}[htbp]
\caption{Complete supervised fine-tuning configuration.}
\label{tab:config}\centering\footnotesize\renewcommand{\arraystretch}{1.15}
\begin{tabularx}{\linewidth}{@{}>{\raggedright\arraybackslash}m{0.28\linewidth}XX@{}}
\toprule
Setting & Phase 1 & Phase 2\\
\midrule
Backbone & \multicolumn{2}{>{\raggedright\arraybackslash}m{0.65\linewidth}@{}}{FLUX.2 [Klein] 4B; bf16 weights}\\
Trainable modules & \multicolumn{2}{>{\raggedright\arraybackslash}m{0.65\linewidth}@{}}{Attention-only LoRA; rank 32, alpha 32; text-stream projections included}\\
Frozen modules & \multicolumn{2}{>{\raggedright\arraybackslash}m{0.65\linewidth}@{}}{Base transformer, MLPs, text encoder, and VAE}\\
Data & \multicolumn{2}{>{\raggedright\arraybackslash}m{0.65\linewidth}@{}}{8,797 change + 13,196 no-change; shuffled; one pass per epoch}\\
Epochs & 0--4 (5 epochs) & 5--9 (5 epochs)\\
Initialization & Pretrained backbone & Own epoch-4 LoRA checkpoint\\
Optimizer state & New AdamW & Reset with \texttt{--lr\_reset}\\
Learning rate & $3\times10^{-5}$ & $5\times10^{-5}$\\
Schedule & 412-step warmup + constant & 300-step warmup--stable--decay\\
Warmup start / decay & Runner default / none & 0.6 / cosine decay ratio 0.15\\
Minimum learning rate & -- & $10^{-5}$ (ratio 0.2)\\
Optimizer & \multicolumn{2}{>{\raggedright\arraybackslash}m{0.65\linewidth}@{}}{AdamW; $\beta=(0.9,0.999)$; $\epsilon=10^{-8}$; weight decay 0.01}\\
Batching & \multicolumn{2}{>{\raggedright\arraybackslash}m{0.65\linewidth}@{}}{8 DDP processes; micro-batch 1; gradient accumulation 2; effective batch 16}\\
Resolution & \multicolumn{2}{>{\raggedright\arraybackslash}m{0.65\linewidth}@{}}{$512\times1024$ (height $\times$ width)}\\
Precision / checkpointing & \multicolumn{2}{>{\raggedright\arraybackslash}m{0.65\linewidth}@{}}{bf16 mixed precision; fp32 optimizer state; gradient checkpointing}\\
Selected checkpoint & -- & Epoch 9\\
\bottomrule
\end{tabularx}
\end{table}

\begin{table}[htbp]
\caption{Complete Flow-GRPO configuration for CrossTimeEdit.}
\label{tab:rlconfig}\centering\footnotesize\renewcommand{\arraystretch}{1.15}
\begin{tabularx}{\linewidth}{@{}>{\raggedright\arraybackslash}m{0.28\linewidth}>{\raggedright\arraybackslash}X@{}}
\toprule
Setting & Value\\
\midrule
Initialization / reference & SFT epoch 9 / frozen SFT epoch 9\\
Trainer / aggregation & Flow-GRPO; DGN, weighted aggregation, then batch normalization\\
Group sampling & 50 conditions per round; 8 candidates per condition; 2 updates per round\\
LoRA / precision & Rank 32, alpha 32; fp32 master weights; bf16 mixed precision\\
Optimizer & AdamW; $7.5\times10^{-6}$; weight decay $10^{-4}$; $\beta=(0.9,0.999)$; $\epsilon=10^{-8}$\\
Gradient norm / RL seed & 1.0 / 42\\
Clipping & Policy ratio $1\pm10^{-3}$; advantage $[-5,5]$; standard-deviation floor $10^{-6}$\\
Reference penalty & Velocity-prediction MSE; coefficient $2\times10^{-3}$\\
Reward weights & IA 0.45; BP 0.30; QP 0.25\\
Sampler & Flow-SDE; noise level 0.7; 10 training / 30 evaluation denoising steps; guidance 1\\
Resolution & $384\times768$ training; $512\times1024$ evaluation\\
Run length / selected epoch & 240 epochs (480 updates) / epoch 200 (400 updates)\\
Reward evaluator & Gemini-3.1-Flash-Lite; independent calls; temperature 0\\
API budget & Initially IA 256 and BP/QP 96 output tokens; all 256 in continuation configs\\
\bottomrule
\end{tabularx}
\end{table}

Validation and test images for the main experiment and all ablation variants were generated using the same inference settings listed in Table~\ref{tab:rlconfig}.

\subsection{Image similarity metrics}
\label{app:metrics}
Generated images are paired with earlier ground-truth panoramas by location ID. The $2048\times1024$ RGB reference is bicubic-resized to $1024\times512$ (width $\times$ height) to match the preselected output. We evaluate each generated panorama against its corresponding earlier ground-truth image. Each metric is computed per image and averaged over the same 1,094 test locations.
\paragraph{PSNR.} For RGB intensities in $[0,1]$, we compute the mean squared error over pixels and channels:
\begin{equation}
\mathrm{MSE}(G,R)=\frac{1}{CHW}\sum_{c,h,w}(G_{chw}-R_{chw})^2,\qquad
\mathrm{PSNR}(G,R)=10\log_{10}\!\left(\frac{1}{\mathrm{MSE}(G,R)}\right).
\label{eq:app_psnr}
\end{equation}
Image-level dB values are averaged; higher is better.

\paragraph{SSIM.} We use \texttt{skimage.metrics.structural\_similarity} with RGB channels (\texttt{channel\_axis=2}), \texttt{data\_range=1.0}, a $7\times7$ uniform window, and the default sample-covariance setting. It compares local luminance, contrast, and structure:
\begin{equation}
\mathrm{SSIM}(G,R)=\frac{(2\mu_G\mu_R+C_1)(2\sigma_{GR}+C_2)}{(\mu_G^2+\mu_R^2+C_1)(\sigma_G^2+\sigma_R^2+C_2)}.
\label{eq:app_ssim}
\end{equation}
Higher values are better. The separately computed Gaussian-window diagnostic is not included in the reported results.

\paragraph{LPIPS.} We use the official LPIPS package with version-0.1 weights and an AlexNet backbone. Let $\hat\phi_{\ell c}$ denote the channel-normalized feature at layer $\ell$ and channel $c$, and $w_{\ell c}$ its learned linear weight:
\begin{equation}
\mathrm{LPIPS}(G,R)=\sum_{\ell}\frac{1}{H_\ell W_\ell}\sum_{h,w,c}w_{\ell c}\bigl(\hat\phi_{\ell c}(G)_{hw}-\hat\phi_{\ell c}(R)_{hw}\bigr)^2.
\label{eq:app_lpips}
\end{equation}
RGB values are mapped from $[0,1]$ to $[-1,1]$ before computing the learned perceptual distance on the complete panorama. Lower values indicate closer perceptual correspondence.

\paragraph{Sharpness Difference (SD).} Following the cross-view evaluation implementation, we compute absolute horizontal and vertical one-pixel differences for generated and reference images, sum the absolute discrepancies between their gradient magnitudes, and exclude the one-pixel border. With $D_aI=|\Delta_aI|$ for $a\in\{x,y\}$, the error and score are
\begin{equation}
\begin{aligned}
S(G,R)&=\sum_{c,h,w}\left(|D_xG-D_xR|+|D_yG-D_yR|\right),\quad e=\frac{128^2S(G,R)}{CHW},\\
\mathrm{SD}(G,R)&=10\log_{10}\!\left(\frac{255^2}{\max(e,0.0001)}\right).
\end{aligned}
\label{eq:app_sd}
\end{equation}
The gradient calculation uses RGB floats in $[0,1]$. These dB scores are averaged; higher scores indicate closer local edge-strength patterns. Absolute values depend on resolution and preprocessing and are comparable only within the shared protocol.

\subsection{Main experiment: reward and validation curves}
\begin{figure}[H]\centering
\includegraphics[width=\linewidth]{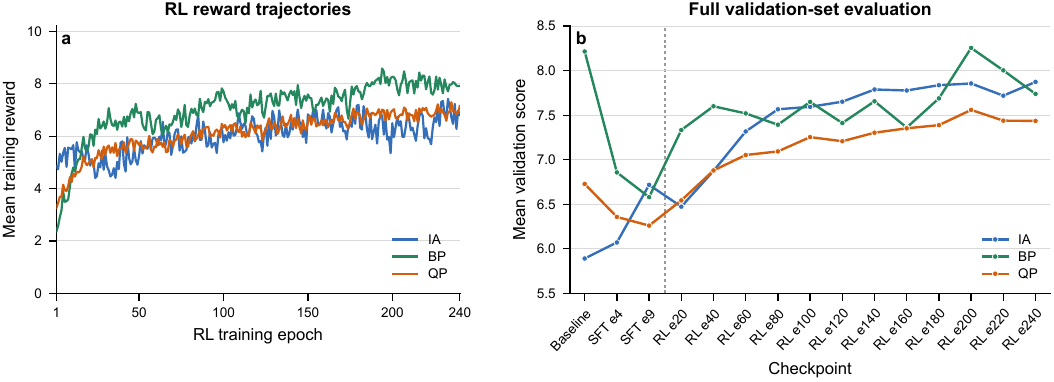}
\caption{Main-experiment training rewards and validation scores. (a) Mean IA, BP, and QP rewards over valid candidates at each RL epoch from 1 to 240. (b) Mean IA, BP, and QP scores on the full 1,095-location change validation split for the pretrained Baseline, SFT checkpoints e4 and e9, and RL checkpoints e20--e240.}
\label{fig:app_main_curves}
\end{figure}

The IA, BP, and QP reward trajectories rose overall from RL epoch 1 to 240, despite noticeable epoch-to-epoch fluctuations (Fig.~\ref{fig:app_main_curves}a).

On the full 1,095-location change validation split, every checkpoint has 1,095 valid scores per dimension, with no failed evaluation requests. Epoch 200 attained the highest weighted Overall among the evaluated RL checkpoints (7.901), with mean scores of 7.856, 8.255, and 7.559 for IA, BP, and QP, respectively. Although IA was marginally higher at epoch 240 (7.873 vs. 7.856), epoch 200 scored higher on BP (8.255 vs. 7.739) and QP (7.559 vs. 7.436), and its weighted Overall exceeded those at epochs 220 (7.734) and 240 (7.723). We therefore selected the epoch-200 checkpoint as the final CrossTimeEdit model (Fig.~\ref{fig:app_main_curves}b).

\subsection{Human preference study}
\label{app:human_preference}
We conduct this study to assess whether the Gemini-3.1-Flash-Lite vision-language model (VLM) produces scores that agree with human preferences. For each model, we define the model-level human win rate as the proportion of pairwise comparisons against other models in which human raters prefer its generated image, with each comparison outcome determined by majority vote. We then compare these win rates with the Overall scores produced by the same VLM under the editing and cross-view evaluation protocols.

\paragraph{Experimental Design.} We invite 15 volunteers with academic backgrounds in geography and related spatial disciplines. We evaluate 11 models through all 55 unordered model pairs, with five comparisons per pair, yielding 275 comparison groups and 50 comparisons per model. Each group randomizes the left/right placement and is assigned to three volunteers for independent judgments. The resulting 825 comparison assignments are randomly shuffled and partitioned into 15 questionnaire sets of 55 items, one for each volunteer. Majority voting determines the winner, and the model-level human win rate is $h_i=W_i/50$. For editing, we compare $h_i$ with the Editing Overall scores of eight models on a 0--10 scale. For cross-view evaluation, we compare $h_i$ with the Cross-view Overall scores of CrossTimeEdit and three cross-view generation models on a 1--5 scale. We also record within-subset human wins over 15 comparisons per cross-view method. Pearson correlation measures score agreement, while Spearman correlation measures ranking agreement using average ranks for ties; the table and figure display tied models with a shared integer rank.

\begin{table}[H]
\caption{Human preference and VLM evaluation results. Human win rates and human ranks use 50 comparisons per model across all 11 models. Editing rank and Cross-view rank denote the rankings by Editing Overall among the eight general image editing models and by Cross-view Overall among the four cross-view generation models, respectively. A dash denotes an inapplicable protocol.}
\label{tab:human_preference}
\centering\footnotesize\setlength{\tabcolsep}{2.5pt}\renewcommand{\arraystretch}{1.12}
\begin{tabularx}{\linewidth}{l|YYYYYY}
\toprule
\textbf{Model} & \textbf{Human win rate (\%)} & \textbf{Human rank (11)} & \textbf{Editing Overall} & \textbf{Editing rank (8)} & \textbf{Cross-view Overall} & \textbf{Cross-view rank (4)}\\
\midrule
Qwen-Image-3.0-Pro & 80\% & 1 & 8.042 & 3 & -- & --\\
GPT-Image-2 & 72\% & 2 & 8.581 & 1 & -- & --\\
Nano Banana 2 Lite & 72\% & 2 & 8.302 & 2 & -- & --\\
CrossTimeEdit & 64\% & 4 & 7.964 & 4 & 3.546 & 1\\
FLUX.2 [Klein] 4B & 58\% & 5 & 6.800 & 7 & -- & --\\
LongCat-Image-Edit & 58\% & 5 & 7.360 & 6 & -- & --\\
FLUX.2 [Klein] 9B & 54\% & 7 & 7.426 & 5 & -- & --\\
OmniGen2 & 48\% & 8 & 4.893 & 8 & -- & --\\
ControlS2S & 30\% & 9 & -- & -- & 2.084 & 2\\
Sat2Density & 14\% & 10 & -- & -- & 2.015 & 3\\
ControlNet & 0\% & 11 & -- & -- & 1.740 & 4\\
\bottomrule
\end{tabularx}
\end{table}

\begin{figure}[H]\centering
\includegraphics[width=\linewidth]{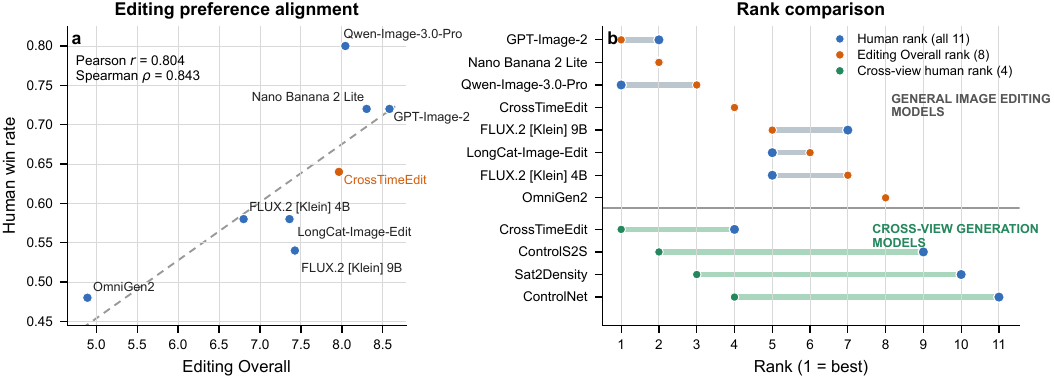}
\caption{Comparison between human preference and VLM evaluation. Left: human win rate versus Editing Overall for the eight general image editing models. Right: human ranks compared with Editing Overall ranks for the eight general image editing models (upper group) and Cross-view Overall ranks for the four cross-view generation models (lower group).}
\label{fig:human_preference}
\end{figure}

\paragraph{Results and Analysis.} The Editing Overall and Cross-view Overall scores reported in Table~\ref{tab:human_preference} are taken from Tables~\ref{tab:main} and~\ref{tab:crossview}, respectively. As shown in Table~\ref{tab:human_preference} and Fig.~\ref{fig:human_preference}, Editing Overall has a positive Pearson correlation with human win rates ($r=0.804$) and a positive Spearman correlation ($\rho=0.843$); 22 of 28 model pairs have concordant orderings. CrossTimeEdit obtains a 64\% human win rate and ranks fourth in both human preference and Editing Overall, while Qwen-Image-3.0-Pro leads human preference and GPT-Image-2 leads Editing Overall. For the four cross-view methods, the human win rates of CrossTimeEdit, ControlS2S, Sat2Density, and ControlNet are 64\%, 30\%, 14\%, and 0\%, respectively, and their Cross-view Overall scores are 3.546, 2.084, 2.015, and 1.740. These values yield Pearson $r=0.896$ and Spearman $\rho=1.000$, indicating a consistent ordering between the cross-view VLM scores and human preferences. The within-subset human wins are 15, 9, 6, and 0 in the same model order, further matching the Cross-view Overall ranking.

The positive correlations in both evaluation settings, together with the consistent model rankings, show that Gemini-3.1-Flash-Lite VLM scores agree with human preference at the model level. These results validate the reasonableness and effectiveness of our VLM-based evaluation protocol for both street-view editing and cross-view generation.
\FloatBarrier

\subsection{Baseline configurations and comparison protocols}
General image editing models use official weights and inference configurations, receiving the same recent street views and local editing instructions.

Cross-view generation models take earlier satellite images as input. We retrained Sat2Density, ControlS2S, and ControlNet on the VIGOR-his training split. Gemini-3.1-Flash-Lite evaluated generated images using CV-C, CV-VR, and CV-PQ on a 1--5 scale; weighted Overall used weights of 0.45, 0.30, and 0.25, respectively. Validation scores were used to select checkpoints for Sat2Density and ControlS2S, whereas we used the final epoch-50 checkpoint for ControlNet. The selected checkpoints were used for test-set inference and evaluation, and the resulting scores are reported in Table~\ref{tab:crossview}.

\paragraph{Sat2Density.} We followed the published hyperparameters, using $256\times256$ satellite images, $512\times128$ panoramas, and 100 sampled points per ray. Trans4PASS provided the sky masks, and illumination was injected using the mean sky histogram computed from the training set. The loss combined $L_1$, $L_2$, Gaussian KL-divergence, feature-matching, perceptual, sky-constraint, and adversarial terms. The generator and discriminator learning rates were both $5\times10^{-5}$. We trained for 30 epochs on one NVIDIA RTX 3090, saving a checkpoint every five epochs; a batch size of 8 with two gradient-accumulation steps gave an effective batch size of 16. We generated validation outputs from seven checkpoints. Epoch 25 achieved the highest weighted Overall. We therefore used epoch 25 for test-set inference.

\paragraph{ControlS2S.} Starting from a pretrained cross-view diffusion model, we trained the GCA module on VIGOR-his using three NVIDIA RTX 3090 GPUs. Training used AdamW with a learning rate of $1\times10^{-6}$, a global effective batch size of 192, and 5,000 effective optimization steps, corresponding to approximately 43.6 epochs. We saved a checkpoint every 500 steps and evaluated ten candidates on the validation set using the three CV criteria. The step-4,500 checkpoint ranked highest and was used for test-set inference.

\paragraph{ControlNet.} We trained ControlNet from Stable Diffusion v1.5 on the VIGOR-his training split using two NVIDIA RTX 3090 GPUs. Training used a global batch size of 4 and a learning rate of $1\times10^{-5}$ for 50 epochs (274,950 optimization steps). We directly selected the final epoch-50 checkpoint for test-set inference and evaluation.

\clearpage
\section{Additional Qualitative Results}

\subsection{VIGOR-his dataset structure and examples}
\label{app:dataset_structure}
\begin{figure}[H]
\centering
\includegraphics[width=\linewidth]{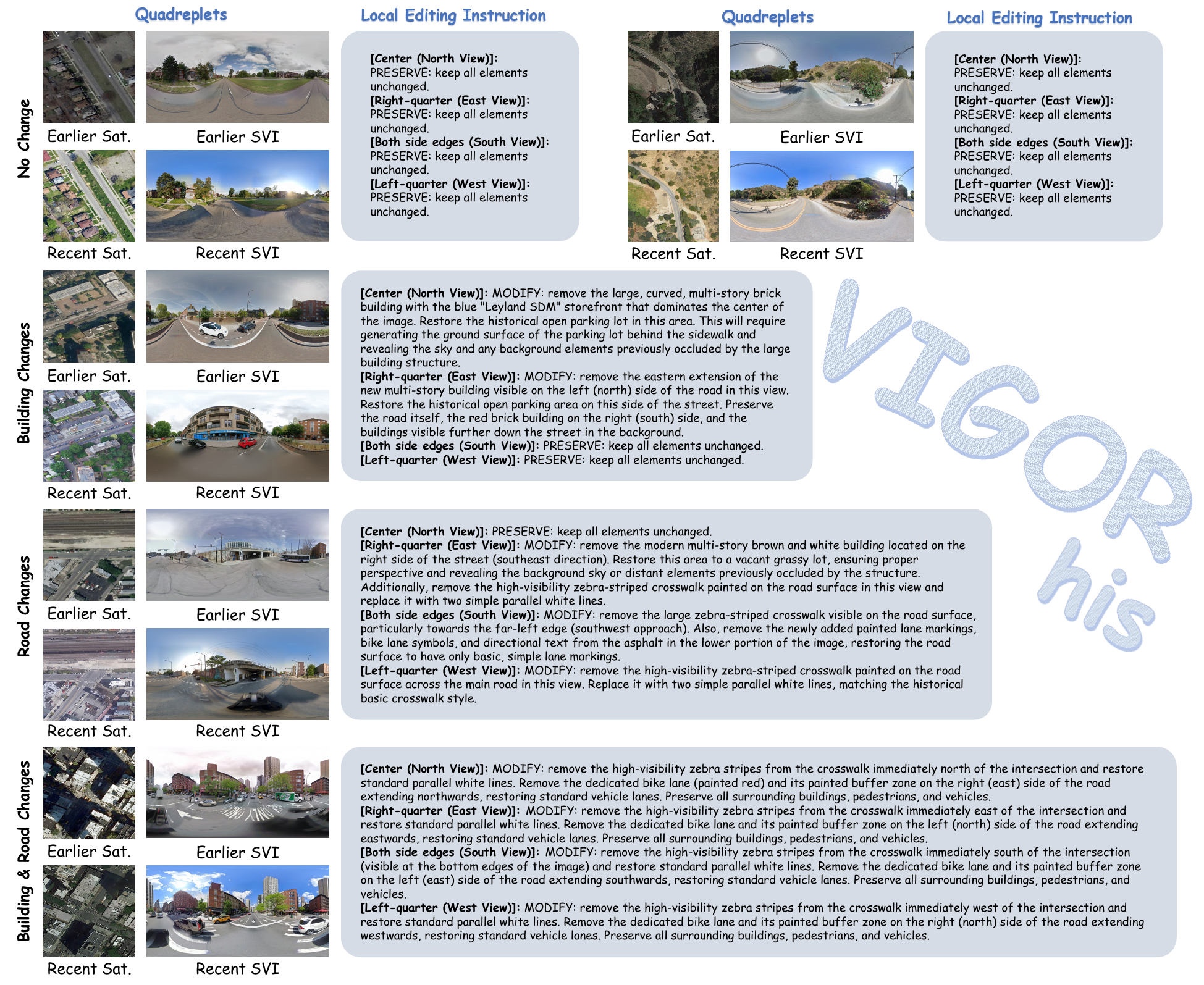}
\caption{Representative VIGOR-his samples and their complete local editing instructions. Each sample contains a temporal cross-view quadruplet of co-located earlier/recent satellite images and street-view panoramas. From top to bottom, the examples cover no change (two locations), building changes, road changes, and combined building and road changes. The instructions organize the panorama by viewing direction and specify region-level `MODIFY' or `PRESERVE' operations.}
\label{fig:app_dataset_structure}
\end{figure}

Fig.~\ref{fig:app_dataset_structure} presents representative VIGOR-his samples spanning no change, building changes, road changes, and their combination. Each sample comprises earlier and recent satellite images and co-located street-view panoramas, forming a temporal cross-view quadruplet. The corresponding satellite-grounded instruction partitions the panorama into the center (north), right quarter (east), both side edges (south), and left quarter (west), and assigns each region either a `MODIFY' operation with the target historical appearance or a `PRESERVE' operation. The no-change examples preserve all four regions, whereas the remaining examples localize building and road modifications while explicitly retaining unchanged scene content. These structured instructions convert temporal cross-view differences into localized supervision for historical street-view generation.

\FloatBarrier

\subsection{Ablation study: qualitative comparison}
\label{app:ablation_qual}
\begin{figure}[htbp]\centering
\includegraphics[width=\linewidth]{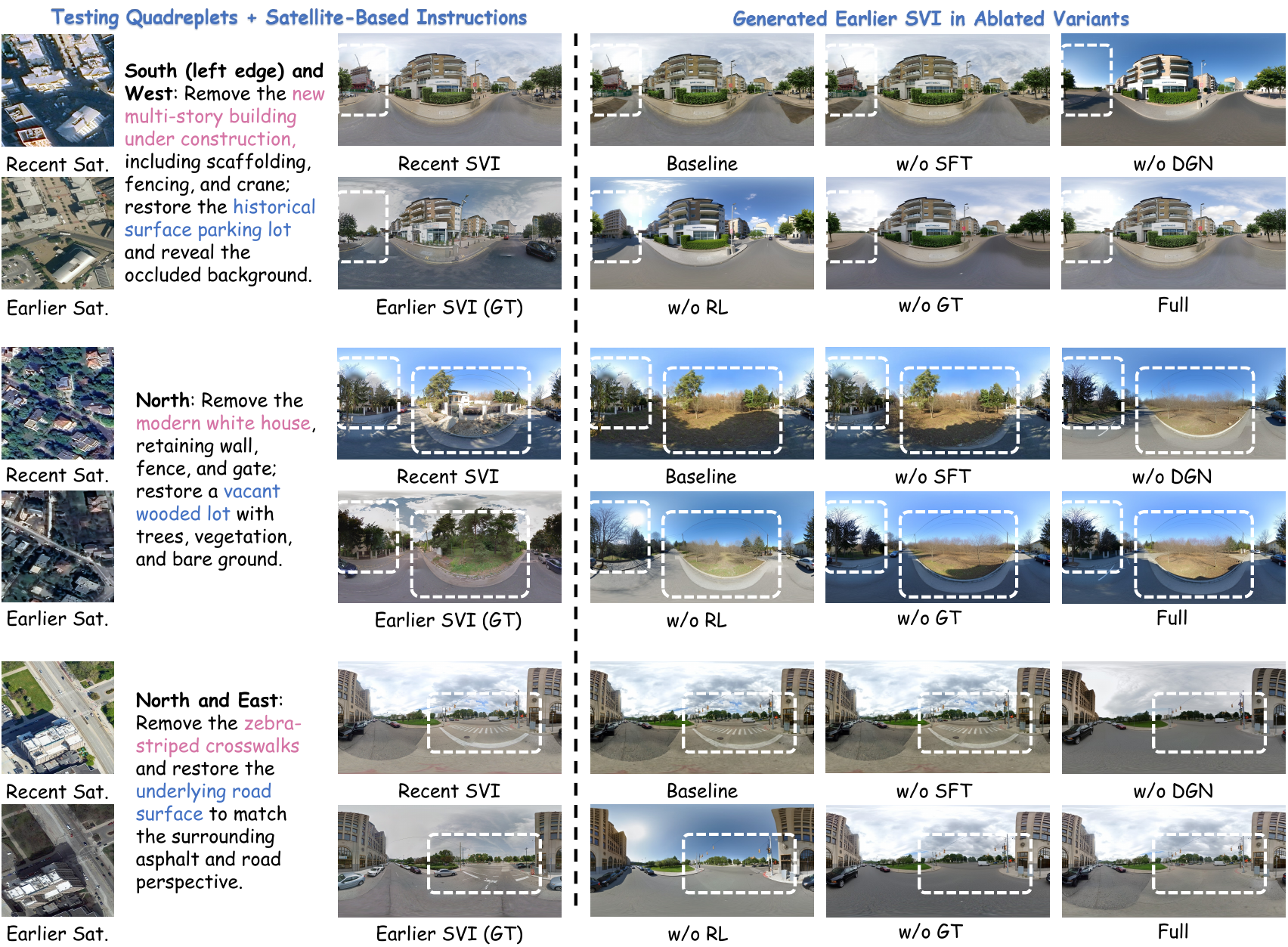}
\caption{Qualitative ablation comparison on three matched locations. Each row shows earlier and recent satellite images, the recent street view, the earlier street-view target, and outputs from the Baseline, w/o SFT, w/o GT, w/o DGN, w/o RL, and Full variants. For readability, the satellite-based instructions shown in the figure are simplified for visualization.}
\label{fig:app_ablation_cases}
\end{figure}

Table~\ref{tab:ablation} and Fig.~\ref{fig:app_ablation_cases} provide complementary quantitative and qualitative evidence for the role of supervised initialization and reward design. The Baseline and w/o SFT variants obtain relatively high background-preservation scores (8.008 and 8.112) but low instruction-alignment scores (5.960 and 6.085), indicating that they tend to retain the input scene while performing the requested historical edits incompletely. After SFT, represented by w/o RL, instruction alignment increases to 6.919, whereas BP and QP decrease to 6.698 and 6.543. This trade-off is also visible in Fig.~\ref{fig:app_ablation_cases}: the SFT-only model more actively restores the requested structures, but introduces larger deviations in non-target regions and lower visual quality than the conservative baseline variants.

The full CrossTimeEdit model further applies RL on top of SFT, increasing IA to 7.950 while restoring and improving BP and QP to 8.264 and 7.631, respectively. The qualitative cases are consistent with this pattern. In the top case, Full better preserves the road and building-surface appearance outside the requested construction-site removal; in the middle case, it retains the building structure on the left while restoring the historical vacant lot; and in the bottom case, it removes the crosswalks while preserving the surrounding road-surface material and layout. The w/o GT and w/o DGN variants achieve instruction-alignment scores close to Full (7.938 and 8.148), but their lower BP and QP scores (7.356/7.508 and 7.433/7.432) are reflected by less reliable preservation of background surfaces and structures in the corresponding cases. These results suggest that SFT provides the capacity for targeted historical editing, whereas the full multi-dimensional RL objective is needed to reconcile edit execution with background preservation and overall visual quality.
\FloatBarrier

\subsection{Qualitative comparison with cross-view generation models}
\label{app:crossview_qual}
\begin{figure}[H]
\centering
\includegraphics[width=\linewidth]{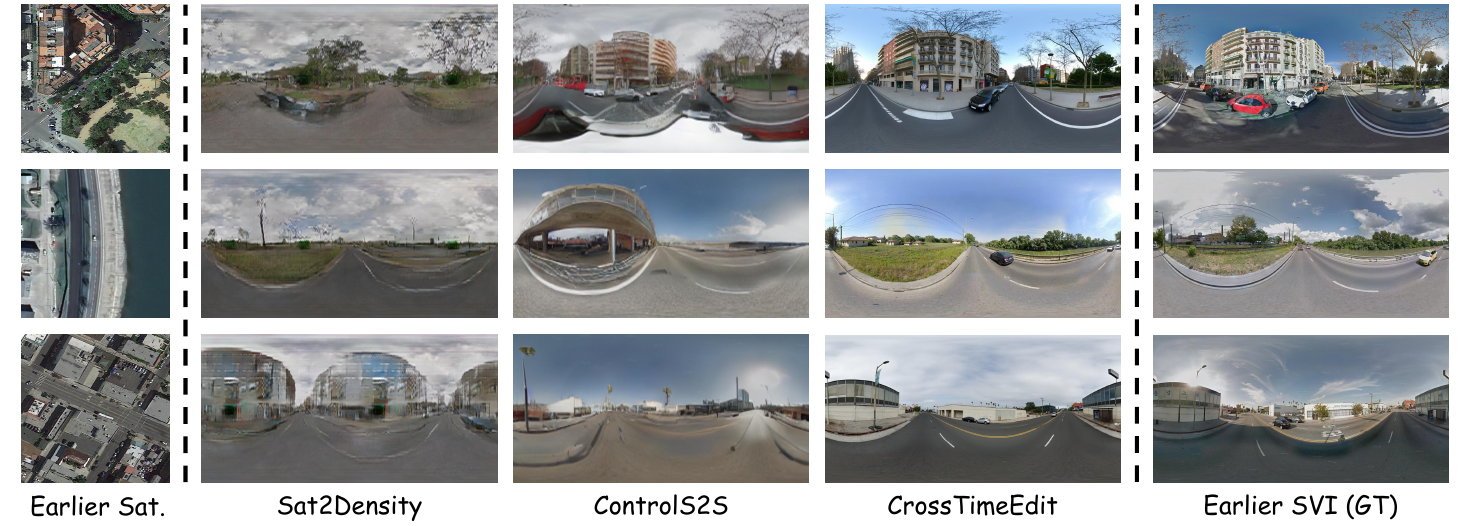}
\caption{Qualitative comparison with cross-view generation models. Each row shows an earlier satellite image, outputs from Sat2Density and ControlS2S, the CrossTimeEdit output, and the corresponding earlier street-view ground truth.}
\label{fig:app_crossview_qual}
\end{figure}

As shown in Fig.~\ref{fig:app_crossview_qual}, the cross-view generation models synthesize street views directly from earlier satellite images, but their outputs differ substantially from the earlier street-view ground truth in viewpoint, scene structure, and appearance. In contrast, CrossTimeEdit uses the recent street view as a visual reference to constrain the viewpoint (location and height) and unchanged appearance, and performs image editing according to satellite-derived change instructions. This formulation better preserves persistent scene content while restoring changed regions toward their earlier states, producing historical street views that more closely match the corresponding earlier observations.

\FloatBarrier
\section{Dataset-construction prompts}
The following prompts retain the complete system and user messages from the source implementations. The screening threshold is applied after scoring: all applicable consistency scores must be at least 7. Score intervals within a prompt are scoring rubrics, not additional retention thresholds. 

\paragraph{D1: Viewpoint screening and change classification (Gemini-3-Flash-Preview).}
\begingroup\fontsize{8}{10}\selectfont
\begin{promptbox}
\begin{verbatim}
# Role You are a professional urban planner and computer vision expert. Your task
is to compare two panoramic street-view images (Image 1 and Image 2) of the same
location from different years and accurately identify real changes in the urban
physical environment.

# Task Execution Logic Silently perform the following two stages of analysis
internally. Do not output any analysis; output only one letter representing the
conclusion at the end.

## Stage 1: Strict viewpoint-alignment screening (mandatory prerequisite) Before
comparing changes, identify large static reference objects in Image 1 (such as
distant buildings, intersection vanishing points, and continuous building
outlines) and locate them in Image
2. Normal parallax (alignment successful): perspective deformation or slight
displacement caused by different capture lanes or a few meters of
forward/backward movement. Reference objects remain in approximately the same
regions. Continue to Stage 2. Major deviation (alignment failed): failed
north-heading correction produces a rotation of tens of degrees. For example, a
building at the exact center of Image 1 moves to the far left/right in Image 2,
or a view facing a building becomes a view facing a straight road. Stop
immediately and output E.

## Stage 2: Change detection (only after successful alignment) If alignment is
adequate, strictly distinguish changes using the following criteria.

Disturbances that MUST be ignored (if only these changes occur, classify as no
change):
1. Climate and lighting: seasonal tree appearance, standing water, cloud cover,
shadow position, and lighting angle.
2. Temporary/mobile objects: pedestrians, vehicles, construction barriers, and
temporary litter.
3. Small ancillary facilities: added or removed streetlights, traffic lights,
signs, fire hydrants, and trash bins.
4. Vegetation and barriers: added or removed trees and bushes; newly built or
removed walls and guardrails. Be extremely careful: never mistake a building
obscured by vegetation, a wall, or a guardrail for an actual change to the
building itself.

Core targets (count only clear changes): Buildings: complete demolition, new
construction, or major renovation of the main structure/exterior facade (such as
adding a second floor to a single-storey building). Ignore minor exterior-wall
color differences. Roads:
1. Major marking updates: large newly added crosswalks, directional arrows, or
road-surface text. Ignore additions, removals, or color changes of ordinary
dashed/solid road lines.
2. Clear material updates: dirt roads converted to asphalt, or conspicuous large
new pavement patches. Ignore minor crack changes, standing water, and reflections
caused by lighting.
3. Road-structure changes: new ramps or branches, or complete road removal.

# Output Format Strictly output exactly ONE uppercase English letter. Do not
include punctuation, spaces, line breaks, or explanation. Choose one of the
following five letters:
E: viewpoint alignment failed with a major deviation.
A: viewpoints aligned, with building changes only.
B: viewpoints aligned, with road changes only.
C: viewpoints aligned, with substantial changes to both buildings and roads.
F: viewpoints aligned, with no changes to the core targets.
\end{verbatim}
\end{promptbox}
\endgroup

The source labels E, A, B, C, and F map to viewpoint mismatch, building change, road change, building-and-road change, and no change, respectively.

\paragraph{D2: Cross-view consistency for changed samples (GPT-5.6-Luna).}
\noindent\textbf{System prompt.}\par
\begingroup\fontsize{8}{10}\selectfont
\begin{promptbox}
\begin{verbatim}
You are a geospatial consistency auditor evaluating the temporal and structural
alignment between top-down satellite orthophotos and 360-degree street-level
panoramas. Your baseline assumption is that discrepancies might exist due to time
mismatches; you must actively hunt for them.

CRITICAL RULE: The street-view images have been pre-rotated so the camera always
faces due North. Vertical reference lines and text labels are explicitly drawn on
the street-view images to guide you:
- Center = North (N): Marked by a thick solid white line and yellow 'N' text.
- Left-quarter = West (W): Marked by a dashed gray line and 'W' text.
- Right-quarter = East (E): Marked by a dashed gray line and 'E' text.
- Both left and right edges = South (S): Marked by dashed blue lines and 'S'
text.

SPATIAL DIRECTION MAPPING: To compare these two different perspectives, you must
understand how their directions align relative to the camera position (the RED
DOT in the satellite image):
- North (N): The center of the street-view corresponds to looking Up (Top) from
the red dot in the satellite image.
- East (E): The right-quarter of the street-view corresponds to looking Right
from the red dot in the satellite image.
- South (S): Both the far-left and far-right edges of the street-view wrap around
to correspond to looking Down (Bottom) from the red dot in the satellite image.
- West (W): The left-quarter of the street-view corresponds to looking Left from
the red dot in the satellite image.

STRICT AUDIT CRITERIA:
1. Blurry Image Rule (FATAL): If the street-view image contains any blurred areas
(whether small patches or large regions), you MUST immediately assign a
[Consistency Score] of 0. Blurriness makes the data unusable for high-quality
training.
2. Building Topology: Check for the presence, absence, relative height, and
footprint shape of permanent buildings in each direction.
3. Land Cover & Developmental State: Evaluate macro-level surface attributes
(e.g., dirt lot vs. paved road, active construction vs. finished building).
4. The "Temporary Object" Rule: Ignore individual mobile objects (e.g., vehicles,
pedestrians, shadows). However, you MUST evaluate permanent infrastructure.
5. Visibility Constraints: Ignore flat rooftop details in the satellite image
that cannot possibly be seen from the ground level.
\end{verbatim}
\end{promptbox}
\endgroup

\noindent\textbf{User prompt.}\par
\begingroup\fontsize{8}{10}\selectfont
\begin{promptbox}
\begin{verbatim}
Analyze the consistency between this street-view and satellite image pair based
on the spatial directional mapping and strict audit criteria. Consider all four
directions (North, East, South, West) in your analysis.

SCORING RUBRIC (0-10):
- 0: [FATAL] The street-view image is blurry (small or large areas), or the
locations are completely disjointed with massive contradictions.
- 1-3: Major mismatches in multiple directions.
- 4-6: Significant temporal changes in one or more directions (e.g., building
under construction vs. completed).
- 7-9: Minor structural differences but overall geometry and land use remain the
same.
- 10: Perfect match. All buildings, road layouts, and macro-level land uses match
perfectly across all directions.

You MUST output ONLY the score on the first line, in this exact format:

[Consistency Score]: <An integer from 0 to 10>
\end{verbatim}
\end{promptbox}
\endgroup

\paragraph{D3: Cross-View and Temporal Consistency Screening for No-Change Samples (GPT-5.6-Luna).}
The no-change branch contains two calls: a four-image audit of cross-view and satellite temporal consistency, and an earlier--recent street-view temporal audit. Their system/user prompt pairs are retained separately below; the four resulting scores must each be at least 7.
\subparagraph{D3(a): Cross-view and satellite temporal audit.}
\noindent\textbf{System prompt.}\par
\begingroup\fontsize{8}{10}\selectfont
\begin{promptbox}
\begin{verbatim}
You are an elite geospatial consistency auditor. You will be provided with ONE
combined image containing FOUR sub-images arranged in a 2x2 grid of the EXACT
SAME geographic location:

TOP-LEFT:    T1 Historical Street View (T1 Street View) — 2:1 panorama with
N/S/W/E direction markers. TOP-RIGHT:   T1 Historical Satellite (T1 Satellite) —
1:1 top-down orthophoto with red dot + yellow north arrow. BOTTOM-LEFT: T2
Current Street View (T2 Street View) — same format as top-left, current time.
BOTTOM-RIGHT: T2 Current Satellite (T2 Satellite) — same format as top-right,
current time.

*NOTE: Each quadrant has its title printed on the top-left corner in a black box.
Quadrants are separated by gray crosshair lines.*

================================================================== PART 1 --
STREET VIEW VISUAL GUIDES (top-left & bottom-left)
================================================================== The
street-view sub-images are 360-degree panoramas flattened into a 2D canvas,
pre-rotated so the camera always faces due North. Vertical reference lines and
text labels:
- Center (thick solid WHITE line + yellow 'N' text) = North
- Left-quarter (dashed GRAY line + 'W' text) = West
- Right-quarter (dashed GRAY line + 'E' text) = East
- Both left and right edges (dashed BLUE lines + 'S' text) = South

================================================================== PART 2 --
SATELLITE VISUAL GUIDES (top-right & bottom-right)
================================================================== Both satellite
sub-images have:
- A RED DOT in the center (exact same camera/focal location).
- A YELLOW ARROW pointing Up (due North).
- N, S, W, E text labels at the top, bottom, left, and right edges.

================================================================== PART 3 --
SPATIAL DIRECTION MAPPING
==================================================================
- North (N): Center of street-view = UP (TOP) from the red dot.
- East (E):  Right-quarter of street-view = RIGHT from the red dot.
- South (S): Far-left and far-right edges wrap = DOWN (BOTTOM) from the red dot.
- West (W):  Left-quarter of street-view = LEFT from the red dot.

================================================================== PART 4 -- YOUR
THREE TASKS ==================================================================

--- TASK A: T1 Cross-View Consistency (top-left vs top-right) --- Evaluate
whether T1 street-view (top-left) matches T1 satellite (top-right). Criteria:

A1. [FATAL BLUR RULE]: If the top-left street-view contains ANY blurred areas,
score MUST be 0. A2. Building Topology: Compare presence, absence, height,
footprint of permanent buildings in all four directions. A3. Land Cover &
Developmental State: dirt lot vs. paved road, construction vs. finished building,
etc. A4. "Temporary Object" Rule: IGNORE vehicles, pedestrians, shadows. DO
evaluate permanent infrastructure. A5. Visibility Constraints: IGNORE flat
rooftop details not visible from ground level.

--- TASK B: T2 Cross-View Consistency (bottom-left vs bottom-right) --- Evaluate
whether T2 street-view (bottom-left) matches T2 satellite (bottom-right). Apply
SAME criteria A1-A5.

--- TASK C: Satellite Temporal Consistency (top-right vs bottom-right) ---
Compare T1 satellite (top-right) vs T2 satellite (bottom-right). Verify
"NO-CHANGE" in physical urban structures.

C1. Focus on Static Structures (CRITICAL): Compare buildings, paved roads, tree
canopies in all four directions from red dot. C2. EXCLUSIONS (MUST IGNORE):
- Image resolution, blurriness, quality differences.
- Color, brightness, contrast, sensor changes.
- Seasonal changes (green vs. brown leaves, snow, shadows).
- Temporary/mobile objects (cars, tents, construction vehicles not altering
permanent structures). C3. "Perfect Match" Rule: If buildings and roads remain
identical, score is 10/10 regardless of color/season/clarity changes.
\end{verbatim}
\end{promptbox}
\endgroup

\noindent\textbf{User prompt.}\par
\begingroup\fontsize{8}{10}\selectfont
\begin{promptbox}
\begin{verbatim}
Conduct the triple-audit on this combined 2x2 grid image. Complete three tasks
one by one.

================================================================== TASK A -- T1
Cross-View Consistency (top-left vs top-right)
================================================================== Evaluate
whether T1 street-view (top-left quadrant) matches T1 satellite (top-right
quadrant). Analyze all four directions (North, East, South, West) relative to red
dot.

SCORING RUBRIC (0-10):
- 10: Perfect structural match. Differences purely
seasonal/color/lighting/quality.
- 7-9: Highly consistent. Very minor structural differences, macro environment
identical.
- 4-6: Noticeable structural changes in one or two directions (new/demolished
building, road altered).
- 1-3: Major mismatches across multiple directions.
- 0: [FATAL] Top-left street-view has blurred areas.

================================================================== TASK B -- T2
Cross-View Consistency (bottom-left vs bottom-right)
================================================================== Evaluate
whether T2 street-view (bottom-left) matches T2 satellite (bottom-right). SAME
criteria as Task A.

SCORING RUBRIC (0-10): same as Task A.
- 0: [FATAL] Bottom-left street-view has blurred areas.

================================================================== TASK C --
Satellite Temporal Consistency (top-right vs bottom-right)
================================================================== Compare T1
satellite (top-right) vs T2 satellite (bottom-right). Verify "NO-CHANGE" in
physical urban structures. Analyze all four directions relative to red dot.

SCORING RUBRIC (0-10):
- 10: Perfect structural match. Differences purely
seasonal/color/lighting/quality.
- 7-9: Highly consistent. Very minor structural differences (tiny shed, one tree
removed), macro environment identical.
- 4-6: Noticeable structural changes in one or two directions.
- 1-3: Major mismatches across multiple directions, large-scale changes.
- 0: Completely disjointed locations with massive contradictions.

================================================================== OUTPUT FORMAT
-- ALL THREE TASKS
================================================================== You MUST
output ONLY the three scores, nothing else, in this exact format:

[T1 Cross-View Score]: <Score 0-10>
[T2 Cross-View Score]: <Score 0-10>
[Satellite Consistency Score]: <Score 0-10>
\end{verbatim}
\end{promptbox}
\endgroup

\subparagraph{D3(b): Street-view temporal audit.}
\noindent\textbf{System prompt.}\par
\begingroup\fontsize{8}{10}\selectfont
\begin{promptbox}
\begin{verbatim}
You are an expert evaluator for a street-view dataset consistency audit.

Two 360-degree street-view panoramas of the SAME geographic location are provided
in a single composite image, one from an EARLIER period and one from a CURRENT
period:
- The TOP half is the HISTORICAL STREET VIEW (earlier period).
- The BOTTOM half is the CURRENT STREET VIEW (current period).

Your task is to verify whether this is a true "NO-CHANGE" location: the permanent
built environment should be structurally identical between the two captures.

Both panoramas are 360-degree images flattened onto a 2D canvas, pre-rotated so
the camera faces due North. Reference lines and labels are drawn on the images:
- CENTER (thick solid white line, yellow 'N') = North
- RIGHT-QUARTER (dashed gray line, 'E') = East
- LEFT-QUARTER (dashed gray line, 'W') = West
- BOTH side EDGES (dashed blue lines, 'S') = South

WHAT COUNTS AS A REAL CHANGE:
- BUILDINGS: any change to a building's static structure (built, demolished,
added/removed storeys, footprint change) OR its facade (cladding, paint
color/pattern, signage, windows, doors) counts as a REAL change.
- ROADS & PAVED SURFACES: changes to lane markings, crosswalks, road-surface
markings, curbs, or the paved ground surface count as a REAL change. IGNORE
surface differences caused purely by lighting, camera shooting style/exposure, or
water ponding on the surface.
- VEGETATION & TREES: changes to trees, hedges, bushes, or grass can be IGNORED —
vegetation is not a structural change.

ALWAYS IGNORE: lighting, weather, parked cars, pedestrians, shadows, seasonal
changes, and minor positional/scale/parallax shifts from different camera capture
points.
\end{verbatim}
\end{promptbox}
\endgroup

\noindent\textbf{User prompt.}\par
\begingroup\fontsize{8}{10}\selectfont
\begin{promptbox}
\begin{verbatim}
Evaluate whether the CURRENT street view (BOTTOM half) is genuinely UNCHANGED
compared to the HISTORICAL street view (TOP half), in terms of the permanent
built environment.

Consider all four directions (N/E/S/W) using the reference lines. Compare
buildings (structure and facade) and roads (surface and markings) between the two
halves. Ignore vegetation changes.

Score SV_T from 0 to 10:
- 10: Buildings and roads are structurally identical between the two street
views; no real change.
- 7-9: Essentially unchanged; only ignorable differences (lighting, weather,
seasonal, parked cars, vegetation).
- 4-6: Noticeable real changes in one or two directions (a building
structure/facade changed, or road markings/surface changed).
- 1-3: Major real changes across multiple directions.
- 0: Completely different scenes / massive redevelopment.

Output ONLY a single line — no reasoning, no explanation, no markdown. Just:
[SV_T]: <0-10>
\end{verbatim}
\end{promptbox}
\endgroup

\paragraph{D4: Satellite-grounded change descriptions (Gemini-3.1-Pro-Preview).}
\noindent\textbf{System prompt.}\par
\begingroup\fontsize{8}{10}\selectfont
\begin{promptbox}
\begin{verbatim}
You are an expert urban geographer and computer vision data annotator. Your task
is to align top-down satellite orthophotos with 360-degree street-level
panoramas. CRITICAL RULE: The street-view images have been pre-rotated so the
camera always faces due North. Vertical reference lines and text labels are
explicitly drawn on the street-view images to guide you: - Center = North (N):
Marked by a thick solid white line and yellow 'N' text. - Left-quarter = West
(W): Marked by a dashed gray line and 'W' text. - Right-quarter = East (E):
Marked by a dashed gray line and 'E' text. - Both left and right edges = South
(S): Marked by dashed blue lines and 'S' text. SPATIAL DIRECTION MAPPING: To
compare these two different perspectives, you must understand how their
directions align relative to the camera position (the RED DOT in the satellite
image): - North (N): The center of the street-view = looking Up (Top) from the
red dot. - East (E): The right-quarter of the street-view = looking Right from
the red dot.
- South (S): Both far-left and far-right edges wrap around = looking Down
(Bottom) from the red dot. - West (W): The left-quarter of the street-view =
looking Left from the red dot. Always describe scenes and map changes by strictly
following this directional mapping. On satellite images: N/S/W/E labels and the
north arrow are drawn in yellow with black outline. You must output exactly in
the requested bracketed formats.
\end{verbatim}
\end{promptbox}
\endgroup

\noindent\textbf{User prompt.}\par
\begingroup\fontsize{8}{10}\selectfont
\begin{promptbox}
\begin{verbatim}
You are given TWO annotated satellite images of the SAME geographic location:
Image 1 — HISTORICAL satellite image (older period) Image 2 — CURRENT satellite
image (recent period)

Each image has a RED DOT at the center (street-view capture point), a YELLOW
ARROW pointing due North (= up), and a small yellow text with black outline in
the top-left corner identifying it as "HISTORICAL satellite image" or "CURRENT
satellite image". Compare both images and identify permanent static structural
changes in the four cardinal directions relative to the red dot.

CRITICAL — GROUND-LEVEL PERSPECTIVE RULE: You must perform mental reconstruction
from the camera position (the RED DOT). Imagine you are standing at the red dot
looking outward in each direction at street level. Describe changes from this
ground-level viewpoint — what would actually be visible to a person standing
there. Do NOT report changes that are only visible from a bird's-eye view and
cannot be seen from the street (e.g., rooftop modifications, solar panel
additions).

WHAT QUALIFIES AS A STRUCTURAL CHANGE (MUST REPORT):
- Buildings: new construction, demolition, footprint expansion/shrinkage.
- Roads: significant changes to road surface material (e.g., dirt→paved) that are
clearly NOT caused by weather or lighting. Addition/removal/reconfiguration of
lane markings, signs, crosswalks, etc. Describe the pre-change structure in
sufficient detail (appearance, size, material, color).

WHAT TO IGNORE (MUST NOT REPORT):
- Seasonal and weather effects: wet vs. dry ground, leaf-on vs. leaf-off
vegetation, shadow differences.
- Vehicles and mobile objects: cars, trucks — regardless of how many or how
different they appear.
- Image quality differences: resolution, blur, color balance, brightness,
contrast between the two periods.

You MUST format your response EXACTLY with these four headers:

[North Direction (Up)]: State what existed historically and what exists now North
of the red dot(Describe the structure in sufficient detail (appearance, size,
material, color)). If unchanged, say "Unchanged: <description>".
[East Direction (Right)]: State what existed historically and what exists now
East of the red dot(Describe the structure in sufficient detail (appearance,
size, material, color)). If unchanged, say "Unchanged: <description>".
[South Direction (Bottom)]: State what existed historically and what exists now
South of the red dot(Describe the structure in sufficient detail (appearance,
size, material, color)). If unchanged, say "Unchanged: <description>".
[West Direction (Left)]: State what existed historically and what exists now West
of the red dot(Describe the structure in sufficient detail (appearance, size,
material, color)). If unchanged, say "Unchanged: <description>".

Keep each section factual, spatially precise, and under 100 words. Total response
under 400 words.
\end{verbatim}
\end{promptbox}
\endgroup

\paragraph{D5: Local editing instruction generation (Gemini-3.1-Pro-Preview).}
\noindent\textbf{System prompt.}\par
\begingroup\fontsize{8}{10}\selectfont
\begin{promptbox}
\begin{verbatim}
You are an expert urban geographer and computer vision data annotator. Your task
is to align top-down satellite orthophotos with 360-degree street-level
panoramas. CRITICAL RULE: The street-view images have been pre-rotated so the
camera always faces due North. Vertical reference lines and text labels are
explicitly drawn on the street-view images to guide you: - Center = North (N):
Marked by a thick solid white line and yellow 'N' text. - Left-quarter = West
(W): Marked by a dashed gray line and 'W' text. - Right-quarter = East (E):
Marked by a dashed gray line and 'E' text. - Both left and right edges = South
(S): Marked by dashed blue lines and 'S' text. SPATIAL DIRECTION MAPPING: To
compare these two different perspectives, you must understand how their
directions align relative to the camera position (the RED DOT in the satellite
image): - North (N): The center of the street-view = looking Up (Top) from the
red dot. - East (E): The right-quarter of the street-view = looking Right from
the red dot.
- South (S): Both far-left and far-right edges wrap around = looking Down
(Bottom) from the red dot. - West (W): The left-quarter of the street-view =
looking Left from the red dot. Always describe scenes and map changes by strictly
following this directional mapping. On satellite images: N/S/W/E labels and the
north arrow are drawn in yellow with black outline. You must output exactly in
the requested bracketed formats.
\end{verbatim}
\end{promptbox}
\endgroup

\noindent\textbf{User prompt.}\par
\begingroup\fontsize{8}{10}\selectfont
\begin{promptbox}
\begin{verbatim}
You are given:
1. A current T2 STREET-VIEW IMAGE (360° panorama, pre-rotated with camera facing
North, with N/S/W/E reference lines drawn).
2. A SATELLITE CHANGE ANALYSIS text describing structural differences between
historical and current satellite images.

SATELLITE CHANGE ANALYSIS: {layer1}

Your task: translate the satellite-confirmed changes into explicit 2D image
editing instructions. CRITICAL: The goal is to guide a Generative AI image
editing model to modify this exact 2D street-view image. You must perform mental
reconstruction — understanding spatial perspective, occlusion, and how each
viewing direction maps to a specific region on the 2D canvas.

STREET-VIEW DIRECTION → 2D CANVAS REGION MAPPING:
- North (N): CENTER of the image (thick solid white line + yellow 'N')
- East (E): RIGHT-QUARTER of the image (dashed gray line + 'E')
- South (S): BOTH the FAR-LEFT and FAR-RIGHT EDGES (dashed blue lines + 'S')
- West (W): LEFT-QUARTER of the image (dashed gray line + 'W')

You MUST output EXACTLY these four headers:

[Center (North View)]: <instruction>
[Right-quarter (East View)]: <instruction>
[Both side edges (South View)]: <instruction>
[Left-quarter (West View)]: <instruction>

Rules:
- Write "PRESERVE: keep all elements unchanged." if that direction has no
satellite-confirmed structural changes.
- Write "MODIFY: remove <current entity>, restore <historical entity>." if
changes occurred. Explicitly instruct how to handle occlusion, reveal
backgrounds, and adjust 3D perspective. Describe the structure in sufficient
detail (appearance, size, material, color).
- If a satellite change is occluded from ground view (e.g., rooftop modification
hidden by building facade), write PRESERVE.
- Be precise about spatial position (e.g., "on the left side of the center view",
"behind the foreground building on the right", "in the southeast direction", "on
the east side of the road").
- Keep each section under 150 words. Total response under 600 words.
\end{verbatim}
\end{promptbox}
\endgroup

\paragraph{D6: Instruction consistency validation (Gemini-3.1-Flash-Lite).}
Each criterion is evaluated in a separate call using the shared system message, its criterion-specific user prompt, and the editing instruction. All three scores must be at least 7; changed samples with all-`PRESERVE' instructions are removed.
The user message also appends \texttt{EDITING INSTRUCTION:} followed by the sample's instruction, then \texttt{COMPOSITE IMAGE (top half = CURRENT STREET VIEW, bottom half = HISTORICAL STREET VIEW):} and the composite image.
\noindent\textbf{Shared system prompt.}\par
\begingroup\fontsize{8}{10}\selectfont
\begin{promptbox}
\begin{verbatim}
You are an expert evaluator for a street-view image editing dataset.

A generative image-editing model is trained to edit the CURRENT STREET VIEW
panorama so that it becomes the HISTORICAL STREET VIEW panorama, guided by a
four-direction editing instruction written from satellite imagery.

- CURRENT STREET VIEW: a 360-degree panorama from the current period. This is the
model's INPUT; the model edits this image.
- HISTORICAL STREET VIEW: a 360-degree panorama from an earlier period. This is
the TARGET / ground-truth the model must reproduce.

The input image is a single COMPOSITE with the two panoramas stacked VERTICALLY,
each labeled at its top-left corner:
- The TOP half is the CURRENT STREET VIEW (model input).
- The BOTTOM half is the HISTORICAL STREET VIEW (target / ground-truth).

Both panoramas are 360-degree images flattened onto a 2D canvas, pre-rotated so
the camera faces due North. Reference lines and labels are drawn on the images:
- CENTER (thick solid white line, yellow 'N') = North
- RIGHT-QUARTER (dashed gray line, 'E') = East
- LEFT-QUARTER (dashed gray line, 'W') = West
- BOTH side EDGES (dashed blue lines, 'S') = South

The instruction has four blocks, one per direction, each marked MODIFY or
PRESERVE:
- "MODIFY: remove <current entity>, restore <historical entity>." — a structural
change happened in this direction.
- "PRESERVE: keep all elements unchanged." — no change happened in this
direction.

The two panoramas may be captured from slightly different camera positions; minor
positional/scale/parallax shifts between the two halves are EXPECTED and should
NOT be penalized. Ignore trivial differences such as lighting, weather, parked
cars, pedestrians, shadows, and seasonal vegetation. Only permanent structures
and ground elements matter.
\end{verbatim}
\end{promptbox}
\endgroup

\noindent\textbf{User prompt: Grounding.}\par
\begingroup\fontsize{8}{10}\selectfont
\begin{promptbox}
\begin{verbatim}
Evaluate ONE dimension: whether the editing instruction is EXECUTABLE on the
CURRENT street view (the TOP half of the composite).

Focus only on the MODIFY blocks. Evaluate EVERY MODIFY block independently. For
each MODIFY instruction, check whether the entity/region it tells the model to
remove is actually present and visible in the CURRENT street view at the stated
direction (use the N/E/S/W reference lines). If a MODIFY target appears in a
DIFFERENT direction than stated, treat it as NOT correctly grounded (major
location error).

The overall G is governed by the WORST-executable MODIFY block — a single
undetectable or misplaced target is a real defect even if other targets are
clear.

Score G from 0 to 10:
- 10: Every MODIFY target is clearly visible in the CURRENT street view at the
correct location; fully executable.
- 7-9: MODIFY targets are visible but with minor imprecision (slight location
drift or vague wording).
- 4-6: Some MODIFY targets are ambiguous, only partially visible, or in the wrong
direction.
- 1-3: Most MODIFY targets are not visible in the CURRENT street view (occluded,
absent, or hallucinated).
- 0: No MODIFY target can be found in the CURRENT street view at all.

Output ONLY a single line — no reasoning, no explanation, no markdown. Just:
[G]: <0-10>
\end{verbatim}
\end{promptbox}
\endgroup

\noindent\textbf{User prompt: Consistency.}\par
\begingroup\fontsize{8}{10}\selectfont
\begin{promptbox}
\begin{verbatim}
Evaluate ONE dimension: whether the HISTORICAL street view (the BOTTOM half of
the composite, the target / ground-truth) matches the MODIFY instructions.

For each MODIFY block, the instruction says "restore <historical entity>".
Evaluate EVERY MODIFY block independently. Check whether that historical entity
is actually present in the HISTORICAL street view at the stated direction, and
whether the HISTORICAL street view visibly differs from the CURRENT street view
in exactly the described way. If the historical entity appears in a DIFFERENT
direction than stated, treat it as a location mismatch.

The overall C is governed by the WORST-matching MODIFY block.

Score C from 0 to 10:
- 10: The HISTORICAL street view clearly shows every historical entity the
instruction says to restore; the instruction matches the ground truth.
- 7-9: It matches with minor discrepancies (extra or missing small objects).
- 4-6: It only partially matches; some described entities are absent, different,
or in the wrong direction in the HISTORICAL street view.
- 1-3: The HISTORICAL street view contradicts the instruction in most MODIFY
regions.
- 0: The HISTORICAL street view shows nothing resembling the historical state the
instruction describes.

Output ONLY a single line — no reasoning, no explanation, no markdown. Just:
[C]: <0-10>
\end{verbatim}
\end{promptbox}
\endgroup

\noindent\textbf{User prompt: Preservation.}\par
\begingroup\fontsize{8}{10}\selectfont
\begin{promptbox}
\begin{verbatim}
Evaluate ONE dimension: whether the regions marked PRESERVE in the instruction
are genuinely unchanged between the CURRENT and HISTORICAL street views.

For each PRESERVE block, compare the corresponding direction region between the
TOP half (CURRENT) and the BOTTOM half (HISTORICAL). Check whether permanent
structures in that region are structurally identical. Evaluate EVERY PRESERVE
block independently — any PRESERVE region with a real structural change drags the
overall P down.

Vegetation rule: a permanently removed or added tree/hedge counts as a structural
change; seasonal leaf/color changes do not.

Score P from 0 to 10:
- 10: All PRESERVE regions are structurally identical between the HISTORICAL and
CURRENT street views.
- 7-9: PRESERVE regions essentially unchanged; only ignorable differences
(seasonal, lighting, parked cars).
- 4-6: One or two PRESERVE regions contain noticeable structural changes the
instruction ignored.
- 1-3: Several PRESERVE regions contain clear structural changes the instruction
missed.
- 0: Major structural changes occur in PRESERVE regions; the instruction missed
them entirely.

If the instruction has NO PRESERVE blocks (all MODIFY), score P =
10.

Output ONLY a single line — no reasoning, no explanation, no markdown. Just:
[P]: <0-10>
\end{verbatim}
\end{promptbox}
\endgroup

\clearpage
\section{Failure-case analysis and limitations}
\label{app:failures}
\begin{wrapfigure}{r}{0.60\linewidth}
\vspace{-0.8\baselineskip}
\centering
\includegraphics[width=\linewidth]{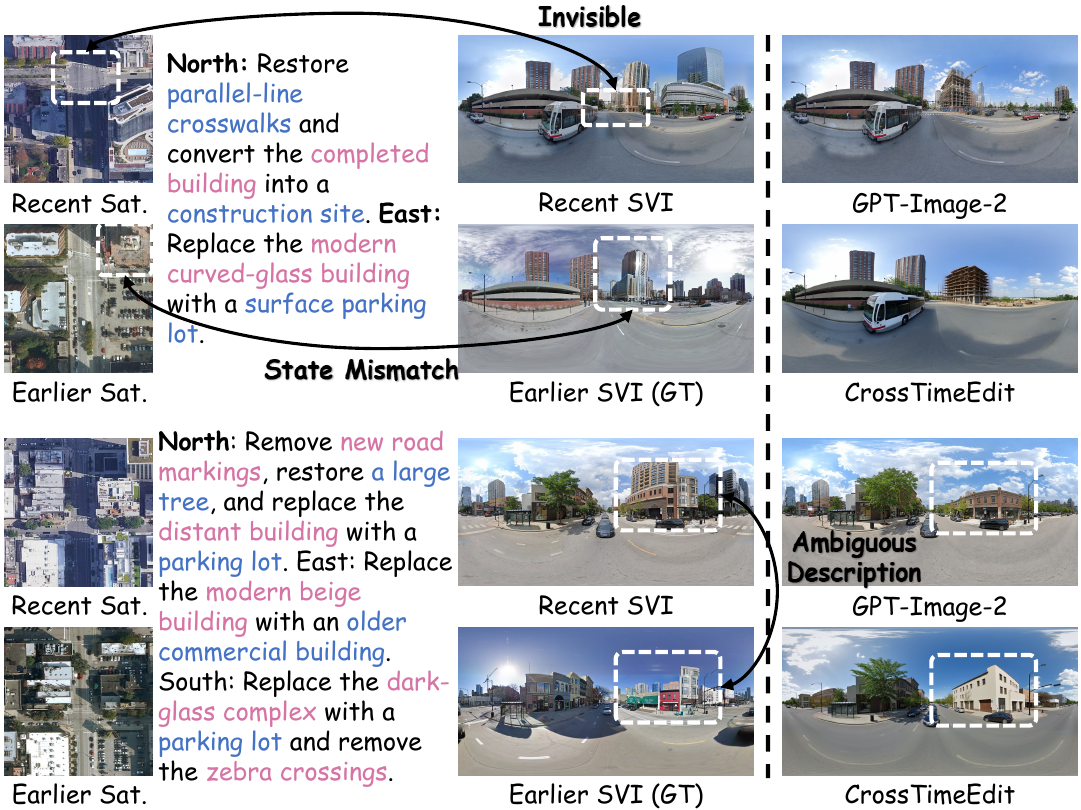}
\caption{Failure cases of historical street-view generation. The top case illustrates limited street-level visibility and a temporal state mismatch within the earlier cross-view pair; the bottom case illustrates ambiguity when generating a complex historical building from limited satellite evidence. For readability, the satellite-based instructions shown in the figure are simplified for visualization.}
\label{fig:app_failures}
\end{wrapfigure}
As shown in Fig.~\ref{fig:app_failures}, the top case exposes two failure sources. First, the parallel-line crosswalks are far from the satellite-image center and are not clearly visible from the corresponding street-view viewpoint, preventing the editing model from reliably executing the instruction. Second, although our dataset-construction pipeline screens the temporal consistency of satellite--street-view pairs from the same period, occasional mismatches remain: here, the earlier satellite image depicts a construction site, whereas the earlier street-view ground truth already contains the completed building. In the bottom case, generating a complex historical building requires detailed facade and structural information that is not available from the satellite view. The resulting satellite-derived instruction is therefore ambiguous, and the generated building can differ substantially from the earlier street-view ground truth.

\paragraph{Limitations.} These cases reveal three principal limitations. First, cross-view visibility differences can make requested changes difficult to observe or verify at street level. This issue also arises under occlusion: for example, a parking lot visible in an earlier satellite image may be hidden behind fences in the corresponding street view, causing the edited result to differ from the actual earlier scene. Second, automatic temporal-consistency screening cannot eliminate every state mismatch between satellite and street-view imagery. Third, satellite observations provide insufficient facade-level evidence for generating complex historical buildings, which can produce underspecified editing instructions and multiple plausible outputs rather than an exact match to the earlier street view.
\FloatBarrier

\section{Future Work}
\label{app:future_work}

\paragraph{Richer historical evidence.} The current framework derives temporal change evidence primarily from paired satellite observations, which cannot fully reveal facades, storefronts, road-level details, or occluded structures. Future work could integrate historical maps, aerial imagery, building footprints, archival photographs, and other ground-level records to reduce ambiguity in satellite-invisible regions and produce more informative local editing instructions.

\paragraph{Independent reward and evaluation models.} CrossTimeEdit currently uses the same VLM protocol to provide RL rewards and evaluate street-view editing results. Future studies could investigate independent reward models, multiple-VLM ensembles, or human-calibrated evaluators to reduce evaluator-specific preferences and limit over-optimization toward a single evaluation protocol.

\paragraph{Bidirectional and multi-temporal generation.} The present formulation generates an earlier street view from a recent street view and temporal satellite evidence. A reverse-time formulation could instead generate present-day street views from historical street views and recent satellite observations. Beyond two acquisition periods, extending the framework to continuous or multi-temporal observations could support modeling longer-term urban evolution rather than a single earlier--recent transition.

\end{document}